%% file: main.tex
\documentclass[journal]{IEEEtran}
\usepackage{ifpdf}
\usepackage{cite}
\usepackage{amsmath}
\usepackage{algorithmicx,algorithm}
\usepackage{array}
\usepackage{fixltx2e}
\usepackage{dblfloatfix}
\usepackage{url}

\usepackage{float}
\usepackage{booktabs}
\usepackage{multirow}
\usepackage{cite}
\usepackage{amsmath}
\usepackage{bbding} 
\usepackage[colorlinks,linkcolor=red,anchorcolor=blue,urlcolor=magenta,citecolor=blue]{hyperref} 
\usepackage[noend]{algpseudocode}

\usepackage{xcolor}

\usepackage{icomma} 
\usepackage{yhmath} 
\usepackage{amssymb} 

\usepackage{xspace}
\makeatletter
\DeclareRobustCommand\onedot{\futurelet\@let@token\@onedot}
\def\@onedot{\ifx\@let@token.\else.\null\fi\xspace}
\def\eg{\textit{e.g\onedot}} 
\def\ie{\textit{i.e\onedot}}

\def\etal{\textit{et al}\onedot}
\makeatother

\usepackage{graphicx}

\begin{document}

\title{PanoFlow: Learning 360$^\circ$ Optical Flow for \\ Surrounding Temporal Understanding}

\author{Hao~Shi†$^{1}$,
        Yifan~Zhou†$^{2}$,
        Kailun~Yang$^{3}$,
        Xiaoting~Yin$^{1}$,
        Ze~Wang$^{1}$,
        Yaozu~Ye$^{1}$,
        Zhe~Yin$^{4}$,
        Shi~Meng$^{5}$,\\
        Peng~Li$^{1}$,
        and~Kaiwei~Wang$^{1}$
\thanks{This work was partly supported by Sunny Optical Technology (Group) Co. Ltd. This work was also supported in part by the National Natural Science Foundation of China (NSFC) under Grant No. 12174341, in part by the Federal Ministry of Labor and Social Affairs (BMAS) through the AccessibleMaps project under Grant 01KM151112, in part by the University of Excellence through the ``KIT Future Fields'' project, and in part by Hangzhou SurImage Technology Company Ltd. \textit{(Corresponding authors: Kaiwei Wang; Kailun Yang.)}}
\thanks{$^{1}$H. Shi,  X. Yin, Z. Wang, Y. Ye, P. Li, and K. Wang are with State Key Laboratory of Modern Optical Instrumentation, Zhejiang University, 310027 Hangzhou, China (email: haoshi@zju.edu.cn; xiaotingyin@zju.edu.cn; wangze0527@zju.edu.cn; yaozuye@zju.edu.cn; lipeng@zju.edu.cn; wangkaiwei@zju.edu.cn).}%
\thanks{$^{2}$Y. Zhou is with Multimedia Laboratory, Nanyang Technological University, 61 Nanyang Dr, Singapore 637335, Singapore (email: yifan.zhou@ntu.edu.sg).}%
\thanks{$^{3}$K. Yang is with Institute for Anthropomatics and Robotics, Karlsruhe Institute of Technology, 76131 Karlsruhe, Germany (email: kailun.yang@kit.edu).}%
\thanks{$^{4}$Z. Yin is with College of Computer Science and Technology, Zhejiang University, 310027 Hangzhou, China (email: pidandan@zju.edu.cn).}%
\thanks{$^{5}$S. Meng is with Data Intelligence Lab, Luokung Technology Corp., 100020 Beijing, China (email: mengshi@luokung.com).}%
}

\markboth{IEEE Transactions on Intelligent Transportation Systems,~2022}%
{Shi \MakeLowercase{\textit{et al.}}: PanoFlow}
\maketitle

\begin{abstract}
\input{Tex_content_red-line/abstract}
\end{abstract}


\IEEEpeerreviewmaketitle

\section{Introduction}

\IEEEPARstart{O}{ptical} flow estimation is one of the fundamental challenges for autonomous driving~\cite{wang2020unsupervised_depth_flow_pose,liu2022camliflow,brebion2021real_flow_event,wang2022_unsupervised_nonocclusion,fang2022traffic_accident_detection}. Flow estimation provides information about the environment and the sensor's motion, leading to a temporal understanding of the world, which is vital for many robotics and vehicular applications, including scene parsing, image-based navigation, visual odometry, and SLAM~\cite{yang2021capturing,gadde2017semantic,hirose2019deep,caruso2015large,zhu2017deep,min2020voldor,teed2021droid}.
With the development of spherical cameras~\cite{gao2022review}, panoramic images are now more easily captured for $360^\circ$ scene perception~\cite{chen2019surrounding_vehicle_detection,liu2022crossmodal_360,petrovai2022semantic_camera}, and can better be integrated with LiDARs due to the similar projection model~\cite{berrio2021camera_lidar}.
However, learning-based methods have always focused on traditional 2D images produced by pinhole projection model based cameras~\cite{sun2018pwc,teed2020raft,shi2022csflow}.
Models designed for a camera with narrow Field-of-View (FoV) are usually sub-optimal for a comprehensive understanding.
Coupling them with $360^\circ$ LiDARs would also directly lead to inherent and domain adaptation problems~\cite{zioulis2018omnidepth}.
Thus, the ability to infer optical flow of a camera's complete surrounding has motivated the study of $360^\circ$ flow estimation.

\begin{figure}[!t]
   \setlength{\abovecaptionskip}{-0.2cm}
   \centering
   \includegraphics[width=1.0\linewidth]{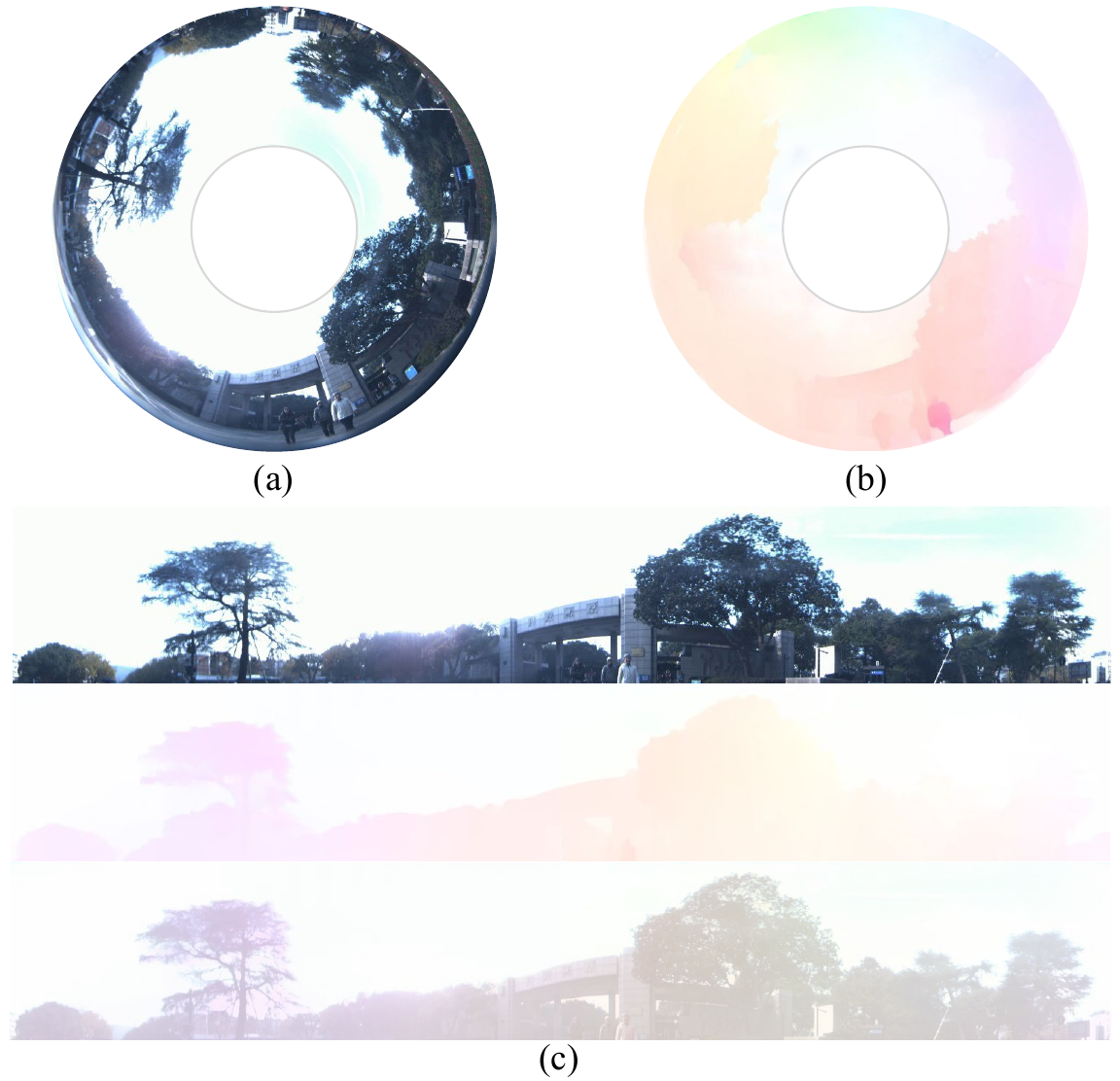} 
   \caption{(a) Raw panoramic annular image captured by our mobile perception system, (b)-(c) the proposed panoramic optical flow estimation on real-world surrounding view for $360^\circ$ seamless scene temporal understanding.}
   \label{fig:introduction_show}
   \vspace{-1.75em}
\end{figure}

Unlike classical linear images, panoramic contents often suffer from severe distortions due to the equirectangular projection (ERP) of spherical cameras~\cite{artizzu2021omniflownet}.
An object will deform to varying degrees at different latitudes in panoramic images, making flow estimation more difficult between the target image and the attended image.
Another critical issue lies in the cyclicity of spherical boundaries, which means there is more than one path from the source point to the target point, and usually there is one shorter and one longer path~\cite{yuan2021360}.
The two routes together form a great circle on the sphere.
In other words, the geometric meanings of the two routes are equivalent. However, traditional learning-based models cannot track pixels moving outside the image boundary, and therefore have no choice but to infer the harder long-distance motion vector, leading to less satisfactory estimation.

To tackle these issues, we introduce a new panoramic flow estimation framework——\textsc{PanoFlow}, to directly estimate dense flow field from panoramic images.
We implement PanoFlow on two different state-of-the-art optical flow networks~\cite{teed2020raft,shi2022csflow} to verify the generality of the proposed framework.
We present the first, to the authors' best knowledge, a \emph{Flow Distortion Augmentation (FDA)} method, that is built on the insight of the distortion induced by ERP, to enhance robustness against deformations in panoramic images.
While distortion augmentation is used in panoramic scene parsing~\cite{yang2019pass,yang2020ds}, it has not been investigated in optical flow estimation, as optical flow is a 2D vector, which incurs further challenges.
Unlike traditional geometric augmentation methods that deal with constant properties, distortion augmentation of optical flow is non-trivial, which has to consider the variation of optical flow because the initial and terminal points of the flow would be distorted to different extent.
By projecting participating images (attended- and target images) and flow ground truth onto the distortion field, we improve the model's ability to generalize to deformed regions.

We put forward two variants of flow distortion augmentation: radial flow distortion (FDA-R) and equirectangular flow distortion (FDA-E). 
Although FDA-E is consistent with the distortion introduced by general ERP, given the smaller FoV of the pinhole dataset, the number of pixels that really participate in supervision is reduced. We therefore also explore the role of FDA-R in overcoming ERP distortion. We found that although their deformed models are not exactly identical, FDA-R also improves the network's ability to handle distorted regions.
From another distortion-adaptive perspective, we further propose to address the distortion by replacing the first layer of the encoder with a deformable convolution layer~\cite{dai2017deformable}.
The proposed FDA and the deformable convolution empower the model to handle characteristic panoramic image distortions and robustify flow estimation.
As a novel data augmentation method, FDA is a plug-and-play module for any learning-based optical flow network.

Furthermore, we give a standard definition of cyclic optical flow suitable for panorama video stream, analyze the properties of cyclic optical flow and compare it with classical optical flow. We then design a \emph{Cyclic Flow Estimation (CFE)} method based on the previous insight to leverage the cyclicity of panoramic images, and convert long-distance estimation to a relatively short-distance estimation. CFE well relieves the stress of the model in large displacement estimation, enabling the model to focus on local fine-grained optical flow estimation. CFE is a general optical flow estimation method and thus can benefit from the advances of narrow-FoV flow estimation methods.
Interestingly, both quantitative and qualitative results show that, compared to the previous best method~\cite{yuan2021360} which estimates on the cubemap plane and the icosahedral tangent plane iteratively, the CFE method is simple, yet very effective.
We also calculate the distribution of the accuracy change with the horizontal FoV before and after using the CFE method, and discover that CFE can significantly improve the optical flow estimation accuracy near the panorama vertical boundary, which is a unique difficulty of panoramic flow estimation.

In addition, to overcome the lack of available panoramic training data and to foster research on $360^\circ$ understanding, we establish and release a new synthetic panoramic flow estimation benchmark of street scenes——\emph{FlowScape}.
We generate the dataset via the CARLA simulator~\cite{dosovitskiy2017carla}. 
FlowScape consists of $6,400$ color images, optical flow, and pixel-level semantic ground truth, providing an environment similar to the real world, thanks to dynamic weather, diverse city street scenes, and different types of vehicles. We use this dataset for learning to infer flow from panoramic content. We also analyze the ground-truth quality of existing optical flow datasets~\cite{artizzu2021omniflownet,yuan2021360,sekkat2022synwoodscape} when only forward optical flow is given, and determine our evaluation datasets according to the observations.

We conduct extensive quantitative experiments on the established FlowScape benchmark.
Compared with the previous best model, the End-Point-Error (EPE) of PanoFlow on this dataset reduces by $27.3\%$.
Further, the EPE of our approach on the public OmniFlowNet dataset~\cite{artizzu2021omniflownet} is reduced by $55.5\%$ compared with the best published results ($3.34$ pixels \textit{vs.} $7.12$ pixels). 
Moreover, a comprehensive set of ablation experiments demonstrates the effectiveness of the proposed FDA and CFE methods.
We additionally conduct qualitative analysis on the public real-world OmniPhotos dataset~\cite{bertel2020omniphotos} to validate our approach in real-world surrounding perception. To further demonstrate the generalization ability of PanoFlow, We also assemble an outdoor data collection vehicle installed with a Panoramic Annular Lens (PAL) system. As shown in Fig.~\ref{fig:introduction_show}, PanoFlow gives sharp and clean omnidirectional optical flow estimation for real-world surrounding scenes. 

In summary, our main contributions are as follows:
\begin{itemize}
   \item We present a rigorous theoretical definition of $360^\circ$ optical flow.
   \item We introduce \emph{flow distortion augmentation}, a new data augmentation method for optical flow networks, which can help models learn to capture the motion cues even on deformed regions.
   \item We propose a generic \emph{cyclic flow estimation} method, which can transform large- to relatively short displacement estimation based on the geometric nature of consecutive panoramas.
   \item We generate \emph{FlowScape}, a new publicly available panoramic dataset that consists of diverse synthetic street scenes, providing both pixel-level flow- and semantic ground truth. We also assess ground-truth quality of existing panoramic flow datasets.
   \item Our entire framework \textsc{PanoFlow} achieves state-of-the-art performance on the established FlowScape benchmark and the public OmniFlowNet dataset.
   \item PanoFlow demonstrates strong generalization ability both on the public real-world OmniPhotos dataset and our captured outdoor panoramic video streams.
\end{itemize}

\begin{table*}[t]
   \centering
   \caption{\textbf{Comparison of existing panoramic datasets for optical flow estimation.}}
   \label{tab:compare_data}
   \centering
   \resizebox{1.0\textwidth}{!}{
   \renewcommand\arraystretch{1.6}{\setlength{\tabcolsep}{0.9mm}{\begin{tabular}{ccccccccc}
            \toprule
            Dataset & Train/Test Split & Groundtruth Quality & ERP format & Resolution & Semantics & Outdoor & Dynamic Weathers & Frames \\
            
            \midrule
            OmniFlowNet Dataset~\cite{artizzu2021omniflownet}                       
            & \XSolidBrush & \textbf{high} & \Checkmark & $384 \times 768$ & \XSolidBrush & \Checkmark & \XSolidBrush & $1500$      \\
            Replica360~\cite{yuan2021360}                       
            & \XSolidBrush & medium/low & \Checkmark & $640 \times 1280$ & \XSolidBrush & \XSolidBrush & \XSolidBrush & $954$     \\
            SynWoodScape~\cite{sekkat2022synwoodscape}                       
            & \XSolidBrush & \textbf{high} & \XSolidBrush & $\textbf{966} {\times} \textbf{1280}$ & \Checkmark & \Checkmark & \XSolidBrush & $500$     \\
            Ours (FlowScape)
            & \Checkmark & \textbf{high} & \Checkmark & $512 \times 1024$ & \Checkmark & \Checkmark & \Checkmark & $\textbf{6400}$     \\
            
            \bottomrule
         \end{tabular}}}}
    \vspace{-1.75em}
\end{table*}

\section{Related work}
\input{Tex_content_red-line/related_work}

\section{PanoFlow: Proposed Framework}
\label{sec:framework}

\subsection{Definition of 360$^\circ$ Optical Flow}
\label{subsec:definition}

\begin{figure}[!t]
   \setlength{\abovecaptionskip}{-0.2cm}
   \centering
   \includegraphics[width=0.9\linewidth]{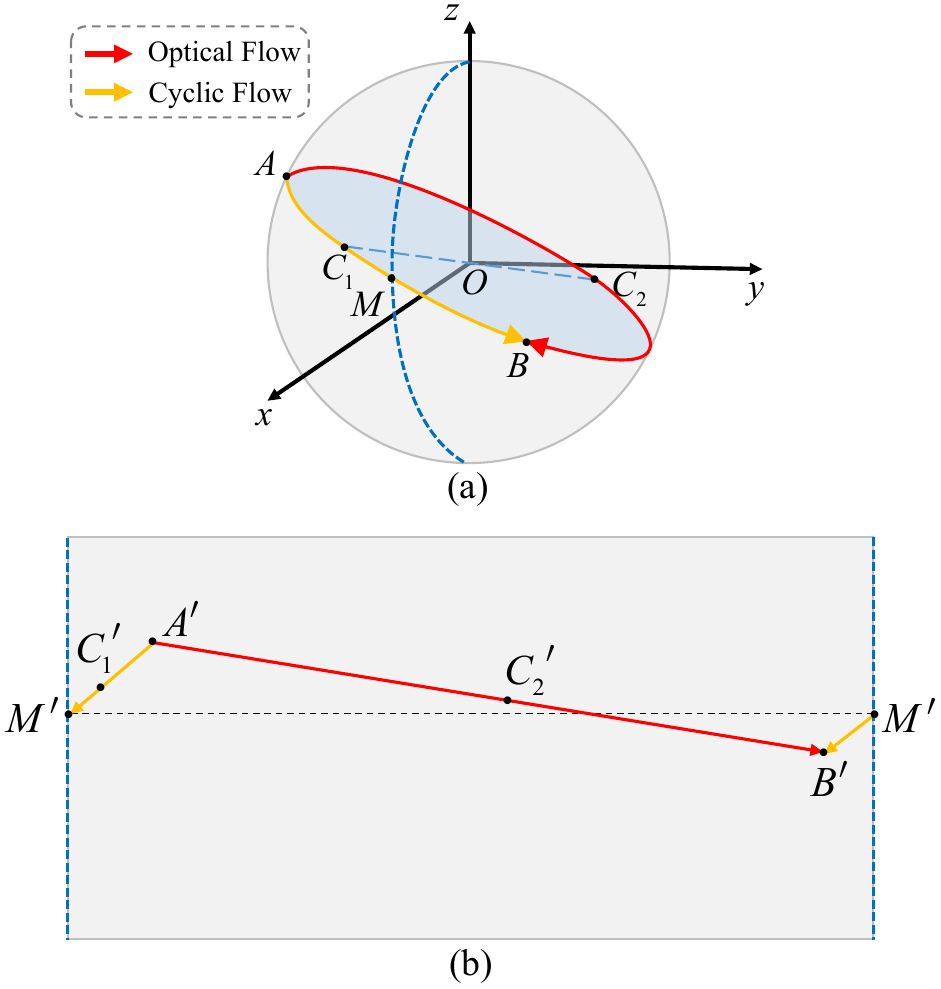}
   \caption{Schematic diagram of cyclic optical flow. (a) The optical flow and the cyclic optical flow form a complementary great circle on the spherical coordinate system, (b) The cyclic optical flow on an equirectangular image has a relatively small displacement and spans the horizontal boundaries of the panorama.}
   \label{fig:define_cyclic}
   \vspace{-1.0em}
\end{figure}

\begin{figure*}[!t]
   \centering
   \includegraphics[width=1.0\linewidth]{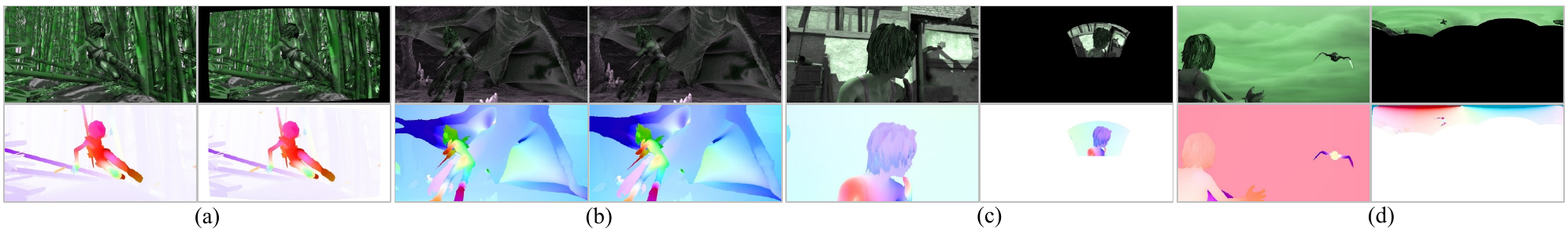}
   \vspace{-2.5em}
   \caption{Visualizations of FDA-R and FDA-E on the Sintel dataset~\cite{butler2012naturalistic}. Notice that the distortion of the image does not affect its intensity, while the color of the optical flow changes with the modulus. (a) Barrel distortion in FDA-R, (b) Pillow distortion in FDA-R, (c) Low latitude distortion in FDA-E, (d) High latitude distortion in FDA-E.}
   \label{fig:FDA}
  \vspace{-1.0em}
\end{figure*}

\begin{figure}[!t]
   \setlength{\abovecaptionskip}{-0.2cm}
   \centering
   \includegraphics[width=1.0\linewidth]{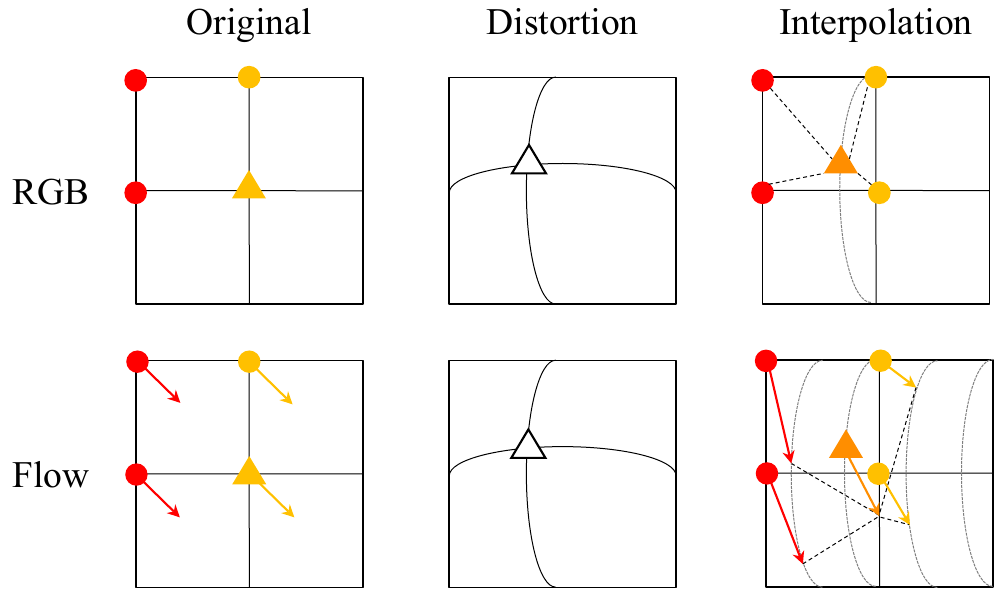}
   \caption{The comparison between RGB image distortion and optical flow distortion. Since the optical flow of grid points has also been modified during distortion, it should be calibrated before interpolation.}
   \label{fig:distortion_comparison}
   \vspace{-1.0em}
\end{figure}

The spherical image does not contain any boundaries, and the coordinates are continuous in any direction on the image~\cite{artizzu2021omniflownet}. However, a boundary parallel to the meridian is naturally introduced in the process of unfolding the spherical image into an equirectangular image as shown in Fig.~\ref{fig:define_cyclic}. Given any point $A$ on the sphere, which moves to another point $B$ after time $t$. Due to the cyclic nature of the sphere itself, there are actually infinite arc trajectories between these two points. Now we only consider two arcs
$\wideparen{AC_{1}B}$
and $\wideparen{AC_{2}B}$ whose range are less than the circumference of the great circle, where $C_1$ and $C_2$ are the vertices of the arcs at both ends, respectively. It is easy to find that these two arcs together form a great circle on the sphere. In the process of spherical unfolding, we can map these two points to $A'$ and $B'$ on the equirectangular image plane $I_e {\in} \mathbb{R}^{H{\times}W}$, respectively, according to the forward ERP:
\begin{equation}
\label{equ:ERP}
    \begin{aligned}
    \left\{
    \begin{array}{rcl}
    x & = & L(\phi-\phi_0)cos\theta_1,\\
    y & = & L(\theta-\theta_0),
    \end{array} \right. 
    \end{aligned}
\end{equation}
where $x$ and $y$ are Cartesian coordinates of the image plane. $\theta {\in} (-\frac{1}{2} \pi, \frac{1}{2} \pi), \phi {\in} (-\pi, \pi)$ are the unit spherical coordinate pitch and yaw, $\phi_0, \theta_0$ are central meridian and central parallel, respectively. $\theta_1$ are the standard parallels. $L$ is the scaling factor.
Therefore, the cyclicity of a great circle on the sphere, is reflected in the cyclicity of the vertical boundary on the equirectangular image:
\begin{equation}
\label{equ:delta_ERP}
    \begin{aligned}
    \delta{x} & = & L cos\theta_1 \cdot (\delta{\phi} \bmod{2 \pi}),
    \end{aligned}
\end{equation}
where $\delta{x}$ and $\delta{\phi}$ denote the variation of Cartesian abscissa and yaw, respectively. $(a \bmod{b})$ indicates $a$ modulo $b$.
Considering a pair of an attended image and a target image, pixels moving out the image boundary on one side will locate to the other side of the image.
Thus, there are two 2D motion vectors that connect the source and target points: one is connected along the interior of the equirectangular image, whereas the other points outside the image boundaries.
These two flow vectors together form a great circle on the spherical image, one shorter and one longer.

For the classical definition of optical flow, given two frames of sequence equirectangular RGB images $I_1$ and $I_2$, we estimate the dense motion vector $(u, v)$ from each pixel $(x, y)$ of $I_1$ to each pixel $(x', y')$ of $I_2$, that is, the optical flow field $\mathbf{V}$, which gives the per-pixel mapping relationship between the source and target.
However, classical optical flow cannot track pixels that move outside the image boundaries, and cannot reflect the boundary circulation of panoramic optical flow. Thus, we define $360^\circ$ optical flow $\mathbf{V_{360}}$ as the shortest path from source to target along the great circle between them, which naturally limits the scalar value of lateral optical flow to $u\leq180^\circ$. For ground-truth flow field $\mathbf{V_{GT}}(\mathbf{x}) = (u, v)$ at pixel index $\mathbf{x}$ of equirectangular images, we can easily convert the optical flow to $360^\circ$ flow:
\begin{equation}
\label{equ:convert}
    \mathbf{V_{360}}(\mathbf{x})=\left\{
    \begin{array}{rcl}
    (u-W, v),  &      & {\frac{1}{2}W < u \leq W};\\
    (u+W, v),  &      & {-W \leq u < -\frac{1}{2}W};\\
    (u, v).    &      & otherwise.
    \end{array} \right. 
\end{equation}
Given a dense cyclic optical flow field $\mathbf{V_{360}}$, we can always find the mapping point $(x', y')$ on $I_2$ from every pixel $(x, y)$ on $I_1$, \ie, cyclic optical flow maintains the temporal continuity of classical optical flow, which can be used to align temporal features when considering boundary cyclicity.

\subsection{Data Augmentation with Flow Distortion}

In optics, distortion is a map projection which makes the straight lines distorted.
Relative to perspective images, the equirectangular transformation can be regarded as a kind of distortion. The models trained on perspective images suffer from the distortion on equirectangular images. To adapt the models to this distortion, we put forward to perform \emph{Flow Distortion Augmentation (FDA)} on the training samples as a novel data augmentation method. 

The distortion of flow is a non-trivial task comparing to general image distortion (Fig.~\ref{fig:distortion_comparison}). For the properties that adhere to the pixels (\eg, RGB or depth), their values would not be modified during distortion. However, the initial and terminal points of optical flow would both be distorted during distortion. To estimate the exact optical flow of a distorted frame, we should calibrate the optical flow of its grid points before interpolation.
Given an undistorted initial point $\mathbf{x_u}=(x_u, y_u)$, the flow field $\mathbf{V_u}$, and a coordinate distortion function $F$ that maps a distorted coordinate to a calibrated coordinate, the calibrated flow field $\mathbf{V_c}$ can be obtained by:
\begin{equation}
\label{equ:generallDistorion}
    \begin{aligned}
    \left\{
    \begin{array}{rcl}
    \mathbf{V_c}(\mathbf{x_d}) & = & F'(\mathbf{x_u} + \mathbf{V_u}(\mathbf{x_u})) - F'(\mathbf{x_u}),\\
    \mathbf{x_d} & = & F'(\mathbf{x_u}),
    \end{array} \right. 
    \end{aligned}
\end{equation}
where $\mathbf{x_d}=(x_d, y_d)$ is the distorted coordinate and $F'$ is the inversion function of $F$.
There are multiple choices for the coordinate distortion function $F$. In this work, we consider radial distortion and equirectangular distortion,
both resulting a remarkable enhancement, which will be discussed in Sec.~\ref{sec:ablation}.
We use the following mapping function  $F_r : F(\mathbf{x_u}) \to \mathbf{x_d}$ to model the radial distortion:
\begin{equation}
\label{equ:RadialDistorion}
    \begin{aligned}
    \left\{
    \begin{array}{rcl}
    x_d & = & {P(r)}(x_c + (x_u - x_c)),\\
    y_d & = & {P(r)}(y_c + (y_u - y_c)),
    \end{array} \right. 
    \end{aligned}
\end{equation}
where $(x_c, y_c)$ is the distortion center (the intermediate point of image by default).
$P(x) {=} x {+} k_2x^2 {+} k_4x^4$ is a polynomial and $r$ is the Euclidean distance from $(x_u, y_u)$ to $(x_c, y_c)$. In practice, we set $k_2{\sim}U( -10^{-6}, 10^{-6}), k_4 {\sim} U(-10^{-14}, 10^{-14})$, which are empirically set and achieve reasonable augmentation effects for images of different resolutions.

For equirectangular distortion $F_e : F(\mathbf{x_u}) \to \mathbf{x_d}$, we transform the coordinates via spherical coordinate system. We first map $\mathbf{x_u}$ on equirectangular image to $\mathbf{x_s}=(x_s, y_s, z_s)$ on unit sphere by: 
\begin{equation}
\label{equ:EquiDistorion1}
    \begin{aligned}
    \left\{
    \begin{array}{rcl}
    x_s & = & sin \frac{\pi y_u}{H} cos \frac{2 \pi x_u}{W},\\
    y_s & = & sin \frac{\pi y_u}{H} sin \frac{2 \pi x_u}{W},\\
    z_s & = & cos \frac{\pi y_u}{H}.
    \end{array} \right. 
    \end{aligned}
\end{equation}
\begin{figure*}[!t]
   \centering
   \includegraphics[width=1.0\linewidth]{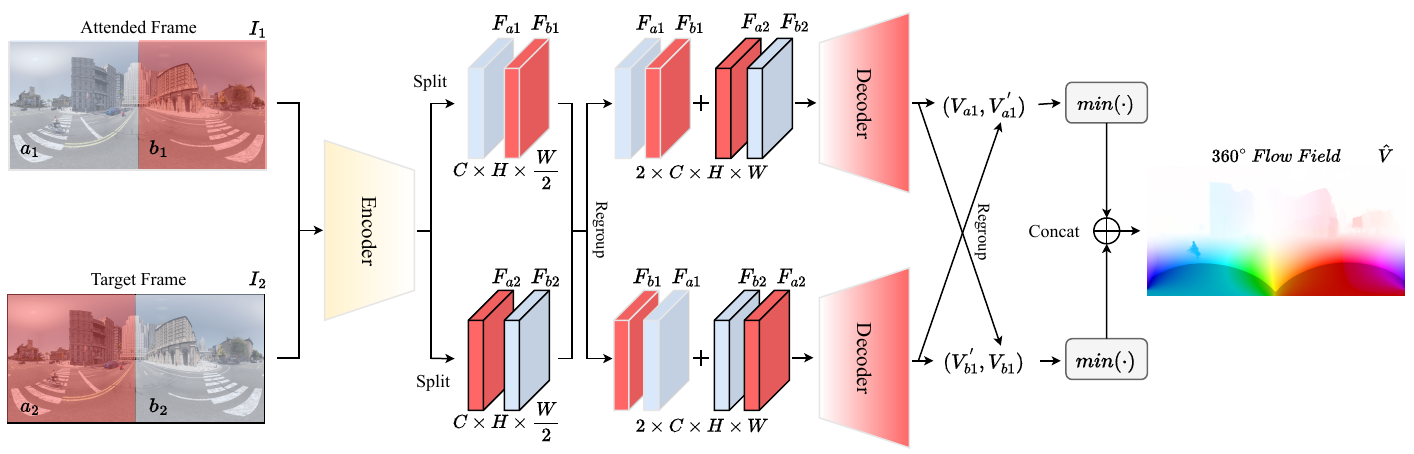}
    \vspace{-2.5em}
   \caption{\emph{Cyclic Flow Estimation}. Partitioned feature maps are extracted from the encoder of the attended frame and the target frame. According to the cyclicity of the left and right boundaries of a panoramic image, the features extracted via the encoder, are regrouped into two feature pairs and sent to the decoder to obtain the complementary optical flow field. The $360^\circ$ flow can finally be obtained via $min(\cdot)$ operations. 
   }
   \label{fig:cyclic_flow_estimation}
   \vspace{-1.5em}
\end{figure*}We then apply random 3D rotation to $\mathbf{x_s}$ by $\mathbf{x_s}^T \gets R_{z}(\theta_z) R_{y}(\theta_y) \mathbf{x_s}^T$, where $R_{y}(\theta_y)$ and $R_{z}(\theta_z)$ are standard 3D rotation matrices about y-axis and z-axis with
$\theta_y {\sim} U(0, \pi)$ and $\theta_z {\sim} U(0, 2\pi)$:
\begin{equation}
\label{equ:EquiDistorion1_R}
    \begin{aligned}
     R_y(\theta_y) = \left[
         \begin{matrix}
           cos{\theta_y} & 0 & sin{\theta_y} \\
           0 & 1 & 0 \\
           -sin{\theta_y} & 0 & cos{\theta_y}
          \end{matrix}
      \right], \\
      R_z(\theta_z) = \left[
         \begin{matrix}
           cos{\theta_z} & -sin{\theta_z} & 0 \\
           sin{\theta_z} & cos{\theta_z} & 0 \\
           0 & 0 & 1
          \end{matrix}
      \right].
    \end{aligned}
\end{equation}
Finally, $\mathbf{x_s}$ is transformed to the perspective coordinate via:
\begin{equation}
\label{equ:EquiDistorion2}
    \begin{aligned}
    \left\{
    \begin{array}{rcl}
    x_d & = & \frac{W}{2tan\frac{\theta_h}{2}}\frac{y_s}{x_s} + \frac{W}{2},\\
    y_d & = & \frac{H}{2tan\frac{\theta_v}{2}}\frac{z_s}{x_s} + \frac{H}{2},\\
    \end{array} \right. 
    \end{aligned}
\end{equation}
where $\theta_h, \theta_v$ are the horizontal and vertical FoV of the perspective image drawn from $U(\frac{\pi}{3}, \frac{2}{3}\pi)$ respectively.

The visualizations of FDA-R and FDA-E are shown in Fig.~\ref{fig:FDA}. Notice that the color of the optical flow changes with the distortion, which is due to that the vector distortion affecting the modulo value of the optical flow. The deformation of FDA-E is homogeneous with that introduced by ERP, while the deformation introduced by FDA-R is radially variable. Taking FDA-E as an example, the position of the color image and optical flow on the spherical surface will change randomly in each iteration instead of being fixed, which improves the robust representation learning ability of the model against distortions. Therefore, the model is able to gradually learn how to handle features with different latitude and longitude on the entire spherical image from the source pinhole data. However, due to the limited FoV of pinhole images, the number of available supervision pixels in FDA-E is actually reduced compared to FDA-R, thus it is necessary to explore the effects of two different optical flow distortion techniques on the distortion robustness of the model. It is verified that FDA improves the adaptation of model by introducing the distorted images to the training data. The ablation experiment of FDA is comprehensively discussed in Sec.~\ref{sec:ablation}.

\subsection{Training with Deformable Receptive Field Encoder}
Unlike pinhole images, equirectangular images suffer from severe geometric distortions in panoramic dense prediction~\cite{zioulis2018omnidepth,yang2019pass}.
While our flow distortion augmentation helps address the deformations from the perspective of training data, classical CNN-based encoders are still limited by the fixed geometry of the convolution kernels, and has insufficient learning ability for deformable features. Therefore, we propose to replace the first convolutional layer of the encoder with deformable convolution~\cite{dai2017deformable} when dealing with $360^\circ$ contents, endowing the model with a more flexible receptive field.
Given a deformable convolution kernel, we extract features at $K$ sampling locations, the weight and grid-specified offset at the k-th location are denoted by $w_k$ and $g_k$, respectively.
In our practice, we replace the feature encoder and context encoder with two deformable convolution layers with a kernel size of $7{\times}7$, thus the kernel is defined with $K{=}49$ and $g_k \in \{ ({-}3,{-}3),({-}3,{-}2),\cdots,(0,0),\cdots,(3,2),(3,3) \}$. 

The distortion-aware features $F_d$ at each position $g_0$ can be obtained via:
\begin{equation}
\label{equ:deformable}
    \begin{aligned}
    F_d(g_0) = \sum\limits^{K}_{k=1} w_k \cdot I(g_0 + g_k + \triangle g_k) \cdot \triangle o_k,
    \end{aligned}
\end{equation}
where $I {\in} \mathbb{R}^{H \times W}$ is the panorama input, $\triangle g_k$ and $\triangle o_k$ are learnable offset and modulation scalar respectively, which are inferred via another convolutional layer:
\begin{equation}
\label{equ:dcn_compensate_1}
    \begin{aligned}
    & \{ \triangle g_k \}_{k=1}^{K} = tanh(\mathcal{C}_{off}(I)[0:2K]), \\
    & \{ \triangle o_k \}_{k=1}^{K} = \sigma( \mathcal{C}_{off}(I)[2K:3K]),
    \end{aligned}
\end{equation}
where $\mathcal{C}_{off}$ is a set of convolutional layers, $[a:b]$ denotes the channel slice from index $a$ to index $b$, $tanh$ and $\sigma$ represent the Tanh and Sigmoid activation function, respectively.
In Sec.~\ref{sec:ablation}, we will show that the use of deformable receptive field encoder further enhances the robustness of the model to distorted images.

\subsection{Inference with Cyclic Flow Estimation}

In order to directly infer $360^\circ$ cyclic flow from equirectangular contents, and relieve the stress of the model in long-distance displacement estimation, we introduce a \emph{Cyclic Flow Estimation (CFE)} method
based on the geometric nature of panoramas.
The structure of CFE is shown in Fig.~\ref{fig:cyclic_flow_estimation}. CFE exploits the cyclicity of the left and right boundaries of equirectangular images, and it is compatible with any optical flow network based on an encoder-decoder structure, \eg, RAFT~\cite{teed2020raft} or CSFlow~\cite{shi2022csflow}. 

\begin{figure*}[t]
   \centering
   \includegraphics[width=0.95\linewidth]{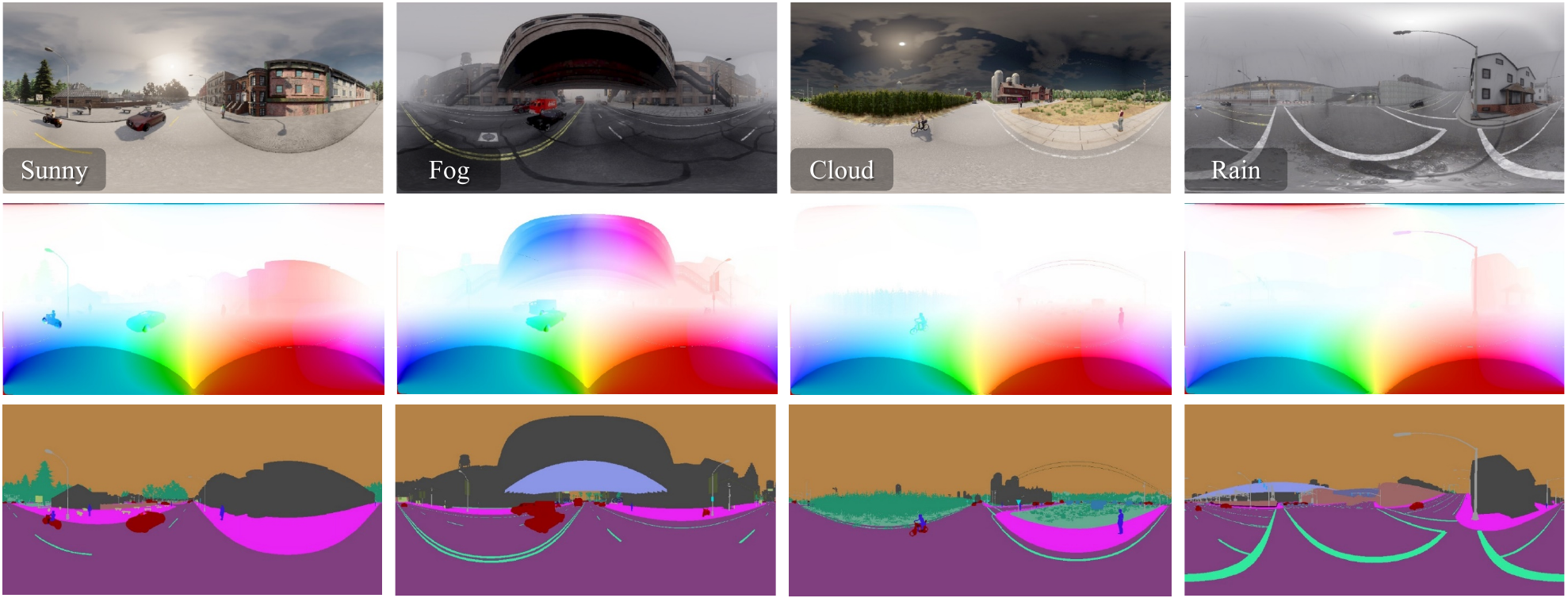}
   \vspace{-1.0em}
   \caption{From up to down: color images, optical flow, and semantics. The proposed \emph{FlowScape} dataset consists of $8$ various city maps in four weathers: sunny, fog, cloud, and rain. We collect $100$ consecutive panoramic images at each random position, resulting in a total of $6,400$ frames with a resolution of $512{\times}1024$~, each with optical flow ground truth and semantic labels, which can be used for training and evaluation. Since the flow field of panoramic images usually contains large displacement that interferes with visualization and fades colors, we modified the visualization method of optical flow based on~\cite{baker2011database}, and lowered the color saturation of optical flow greater than the threshold. Please refer to our open-source documentation for details.}
   \label{fig:dataset}
   \vspace{-1.5em}
\end{figure*}

Specifically, we first use a convolutional network as the encoder $e(\cdot)$ to extract features $F_1, F_2 {\in} \mathbb{R} ^ {C \times H \times W}$ from the input two frames of equirectangular images $I_{1}, I_{2} {\in} \mathbb{R} ^ {3 \times h \times w}$.
Then, the features are split along the horizontal centerline into $F_{a1}, F_{b1} {\in} \mathbb{R} ^ {C \times H \times \frac{W}{2}}$ and $F_{a2}, F_{b2} {\in} \mathbb{R} ^ {C \times H \times \frac{W}{2}}$, respectively.
We regard the process of feature encoding as rigid, that means, we should obtain exactly the same features for the same image input.
Therefore, when swapping the left and right regions of the input image, the resulting feature maps should also be approximately left-right swapped.
Based on the above observations, we can regroup the feature maps as two feature pairs $P_1, P_2 {\in} \mathbb{R} ^ {2 \times C \times H \times W}$:
\begin{equation}
\label{equ:regroup_feature}
   \begin{split}
      \left\{
      \begin{aligned}
         P_1 & = & \{ F_{a1} \oplus F_{b1}, F_{a2} \oplus F_{b2} \}, \\
         P_2 & = & \{ F_{b1} \oplus F_{a1}, F_{b2} \oplus F_{a2} \}.
      \end{aligned}
      \right.
   \end{split}
\end{equation}
where $\oplus$ means a concatenate operation. Since the RAFT structure~\cite{teed2020raft} contains an additional context encoder $c(\cdot)$, the context feature maps $C_{a1}, C_{b1} {\in} \mathbb{R} ^ {C \times H \times \frac{W}{2}}$ extracted from $I_1$ should also be regrouped into $P_{c1}, P_{c2} {\in} \mathbb{R} ^ {C \times H \times W}$:
\begin{equation}
\label{equ:regroup_context}
   \begin{split}
      \left\{
      \begin{aligned}
         P_{c1} & = & \{ C_{a1} \oplus C_{b1}\}, \\
         P_{c2} & = & \{ C_{b1} \oplus C_{a1}\}.
      \end{aligned}
      \right.
   \end{split}
\end{equation}
We then stack the feature pairs with context respectively, which will be further sent to the decoder $d(\cdot)$. Subsequently, the decoder will estimate two flow fields $\mathbf{V},~ \mathbf{V^{'}} {\in} ~ \mathbb{R} ^ {2 \times h \times w}$. 
\newpage
\noindent The flow estimations are split along the horizontal centerline into $\mathbf{V_{a1}}, \mathbf{V_{b1}} {\in} \mathbb{R} ^ {2 \times h \times \frac{w}{2}}$ and $\mathbf{V^{'}_{a1}}, \mathbf{V^{'}_{b1}} {\in} \mathbb{R} ^ {2 \times h \times \frac{w}{2}}$. Assuming that the estimation is unbiased, for any pixel $(x, y)$ in area $a$, we consider that $\mathbf{V_{a1}}(x, y)$ and $\mathbf{V^{'}_{a1}}(x, y)$ form a pair of complementary optical flows end to end, and these two 2D motion vectors together form a great circle on the sphere. The same is true for area $b$. According to our definition of $360^\circ$ optical flow, the final $360^\circ$ flow field $\mathbf{\hat{V}}$ is obtained:
\begin{equation}
\label{equ:final_flow}
    \mathbf{\hat{V}} = min(\mathbf{V_{a1}}, \mathbf{V^{'}_{a1}}) \oplus min(\mathbf{V^{'}_{b1}}, \mathbf{V_{b1}}).
\end{equation}
We emphasize again that CFE is a generic flow estimation method based on the assumption that the encoding process should be rigid, which can replace the large displacement estimation with the small displacement estimation when dealing with panoramic contents. According to our analysis of the geometric nature of consecutive panoramic frames in Sec.~\ref{subsec:definition}, CFE is able to cope with the intrinsically most difficult part of long-range cyclic estimation in panoramic optical flow, without having to estimate dozens of times on the tangent plane of the regular polyhedron like the previous method~\cite{yuan2021360}. Considering that large displacement estimation is much more challenging for the model, CFE can significantly enhance the prediction reliability. With the proposed CFE method, we eliminate redundant encoding calculations, and ensure computational efficiency while accurately estimating $360^\circ$ optical flow. Another naive idea is to use circular convolutions to replace classical convolutional layers. However, we will show in ablation studies (Sec.~\ref{sec:ablation}) that this method only has a limited circularity for cyclic flow, and thus it is not suitable for panoramic flow estimation.

\section{FlowScape: Established Synthetic Dataset}
\label{sec:FlowScape}

End-to-end learning of deep neural networks requires a large amount of annotated ground truth data.
Although for pinhole cameras this can be partly resolved by using scanning LiDARs and multiple sensors~\cite{geiger2013vision,kondermann2016hci}, such an approach is unpractical for $360^\circ$ images, considering that panoramic camera and LiDAR will block each other and have a large divergence in their resolutions. In addition, the point cloud data given by LiDARs is sparse, thus it is difficult to obtain dense ground-truth values of optical flow in the real world.
Even when these flaws are patched using algorithms during acquisition, additional errors are still introduced.
On the other hand, synthetic datasets are popular for learning flow estimation due to the lack of real-world training data~\cite{butler2012naturalistic,dosovitskiy2015flownet,mayer2016large}. Extensive investigations have demonstrated that generalization from synthetic- to real scenes is feasible for optical flow tasks~\cite{teed2020raft,xu2021high,shi2022csflow}. 

\noindent\emph{\textbf{FlowScape dataset:}}
We notice that there is a lack of an open panoramic optical flow dataset that can be used for training and credible numerical evaluation.
Therefore, we advocate to generate a dataset with ground-truth flow by synthesizing both the color image and flow via the CARLA simulator~\cite{dosovitskiy2017carla}. Specifically, we use eight open-source maps given by CARLA.Our virtual collection vehicle contains $6$ pinhole color cameras, $6$ pinhole optical flow cameras, and $6$ pinhole semantic cameras, all of which have a FoV of $90^\circ{\times}90^\circ$, are in the same spatial viewpoint and keep synchronized timestamps. Taking color images as an example, six orthogonal viewing angles are obtained to form a cubemap panorama $\{I_f, I_r, I_b, I_l, I_u, I_d\} {\in} \mathbb{R} ^{h \times w}$, including front, right, back, left, top, and bottom view. We can then acquire the equirectangular image $I_e {\in} \mathbb{R} ^{H \times W}$ with a FoV of $180^\circ{\times}360^\circ$ by using a cubemap-to-equirectangular algorithm as a post-processing. Given four horizontally views $\{I_f, I_r, I_b, I_l\}$ of the cubemap format, we can calculate their corresponding coordinates $(x, y)$ on the equirectangular image plane:
\begin{equation}\label{eq:C2E_horizontal}
    \begin{aligned}
    \left\{
    \begin{array}{rcl}
    x & = & \frac{W}{2} \cdot tan(\phi - m\frac{\pi}{2}),\\
    y & = & -\frac{H}{2} \cdot \frac{tan\theta}{cos(\phi - m\frac{\pi}{2})},
    \end{array} \right. 
    \end{aligned}
\end{equation}
where the view index $m{=}\{1,2,3,4\}$, $\theta {\in} (-\frac{1}{2} \pi, \frac{1}{2} \pi)$, $\phi {\in} (-\pi, \pi)$ are the angular coordinates. For the upper and lower views $\{I_u, I_d\}$:
\begin{equation}\label{eq:C2E_ud}
    \begin{aligned}
    \left\{
    \begin{array}{rcl}
    x & = & \frac{W}{2} \cdot tan(\frac{\pi}{2} - \theta) sin(\phi),\\
    y & = & \frac{H}{2} \cdot tan(\frac{\pi}{2} - \theta) cos(\phi + n\pi),\\
    \end{array} \right. 
    \end{aligned}
\end{equation}
where the view index $n{=}\{0,1\}$. Given $6$ views in the cubumap-format panorama with a resolution of $1024{\times}1024$, the reprojection latency of the panoramic image is $0.37s$, and the panoramic optical flow takes $1.29s$, both test on the Intel i5-12600K CPU with Python 3.9 implementation.

\begin{figure}[t]
   \centering
   \includegraphics[width=1.0\linewidth]{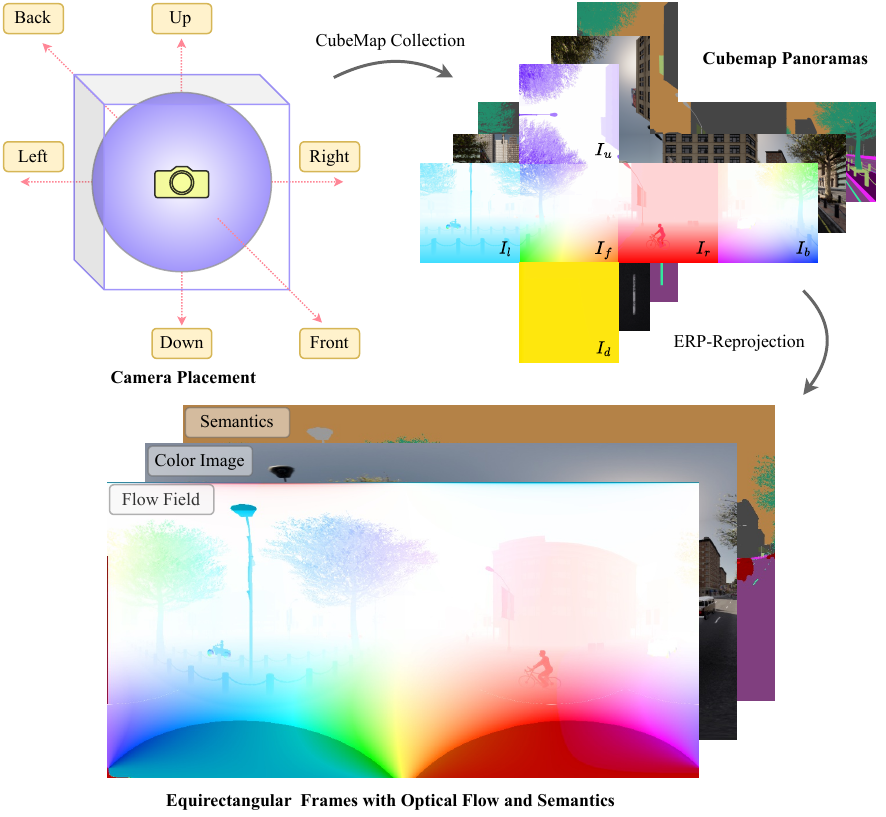}
   \vspace{-1.0em}
   \caption{Schematic diagram of virtual pinhole cameras placement. The six virtual cameras are located at the same spatial viewpoint, and their orientations are orthogonal to each other. The collected cubemap panoramas are reprojected by ERP to obtain complete field-of-view equirectangular frames with dense flow filed and semantics.}
   \label{fig:camera_placement}
  \vspace{-1em}
\end{figure}

We set $100{\sim}120$ initial collection points on each of the $8$ open source maps, and all of them are on the road. During collection, a tracing renderer is used to render our dataset by placing these pinhole cameras at a starting position $P \in \mathbb{R} ^ {3}$ in the scene, which is randomly sampled from the initial collection points of the map.
For each map, we augment the dataset by changing weather, including \emph{sunny} ($62.5\%$), \emph{cloud} ($12.5\%$), \emph{fog} ($12.5\%$), and \emph{rain} ($12.5\%$) to form FlowScape and assess the robustness of optical flow estimation in various conditions. During the collection process, as the number of vehicles and pedestrians increases, the rendering overhead will increase slightly. For purpose of controlling the stability of the data collection while maintaining the richness of the foreground, we set the generation upper limit of vehicles and pedestrians to $200$ for all the maps.
In order to ensure a good diversity of the synthetic data, we only gather $100$ frames with a frame rate of $30Hz$ for each position.
Considering the unreliable optical flow at infinity such as that on points in the sky, we additionally provide ground-truth values of pixel-wise semantic segmentation for selection, which could also be beneficial for panoramic semantic understanding tasks~\cite{yang2019pass,yang2020ds}.
The semantic labels follow CARLA's setting (Fig.~\ref{fig:dataset}).
Overall, FlowScape provides $6,400$ panoramic images of diverse street scenes, each with ground truth of both optical flow and semantic labels.

\noindent\textbf{Photoconsistency analysis:}
As shown in Tab.~\ref{tab:compare_data}, we compare the existing panoramic optical flow datasets. SynWoodScape~\cite{sekkat2022synwoodscape}, OmniFlowNet~\cite{artizzu2021omniflownet}, and Replica360~\cite{yuan2021360} are three small datasets for evaluating panoramic optical flow. Due to their small size, they are not suitable for the training of neural network based methods. We further explore the photoconsistency~\cite{baker2011database} by introducing photometric error (PE) and warped photometric error (WPE) to evaluate the ground-truth optical flow quality of the dataset when only forward flow is given:
\begin{equation}
\label{equ:warp_error_1}
    PE = \frac{1}{HW} \sum\limits_{\mathbf{x}} \lvert I_1(\mathbf{x}) - I_2(\mathbf{x}) \rvert,
\end{equation}
\begin{equation}
\label{equ:warp_error_2}
    WPE = \frac{1}{HW} \sum\limits_{\mathbf{x}} \lvert I_1(\mathbf{x} + \mathbf{V_{GT}(\mathbf{x})}) - I_2(\mathbf{x}) \rvert,
\end{equation}
where $\mathbf{x}$ is the pixel index and $\mathbf{V_{GT}}$ is the ground-truth flow field. Obviously, the quality of the ground-truth flow is high when WPE is significantly lower than PE, \ie, a high-quality optical flow field can convert one image to the next as much as possible~\cite{mccane2001benchmarking}. 
We consider forward optical flow that results in a reduction in interpolation error of less than $10\%$ to be medium/low-quality ground truth.
We perform ground-truth quality analysis on the popular perspective optical flow dataset~\cite{dosovitskiy2015flownet, mayer2016large, butler2012naturalistic} and these three panorama datasets separately, and the results are shown in Tab.~\ref{tab:gt_analy}.
Compared to the public OmniFlowNet and our FlowScape datasets, the ground-truth flow of Replica360 dataset seems to be unreliable. Consequently, our quantitative evaluations are performed on the first two datasets.

\begin{table}[t]
   \caption{\textbf{Ground-truth quality analysis of optical flow datasets.}}
   \label{tab:gt_analy}
   \resizebox{0.48\textwidth}{!}{
   \centering
   \renewcommand\arraystretch{1.4}{\setlength{\tabcolsep}{1mm}{\begin{tabular}{cccccc}
            \toprule
            Modal& Dataset & PE & WPE & Diff. & GT Quality
            \\

            \midrule
            \multirow{3}{*}{Perspective} 
            & FlyingChairs~\cite{dosovitskiy2015flownet} & 23.35 & 12.43 & $\uparrow46.8\%$ & high \\
            & FlyingThings~\cite{mayer2016large} & 25.57 & 19.26 & $\uparrow24.7\%$ & high \\
            & MPI-Sintel~\cite{butler2012naturalistic} & 13.93 & 7.12 & $\uparrow48.9\%$ & high \\
            
            \midrule
            \multirow{4}{*}{Panorama} 
            & SynWoodScape~\cite{sekkat2022synwoodscape} & 10.22 & 7.20 & $\uparrow29.5\%$ & high \\
            & OmniFlowNet~\cite{artizzu2021omniflownet} & 8.47 & 5.75 & $\uparrow32.1\%$ & high \\
            & Replica360~\cite{yuan2021360} & 19.76 & 18.63 & $\uparrow5.7\%$ & medium/low \\
            & Ours (FlowScape) & 5.96 & 3.07 & $\uparrow48.7\%$ & high \\
            
            \bottomrule
         \end{tabular}}}}
\end{table}

\section{Experiments}
\label{sec:experiments}
We conduct experiments using two typical learning-based flow method~\cite{teed2020raft,shi2022csflow} to verify the proposed PanoFlow framework. We confirm the role of the key components in PanoFlow through ablation experiments.
For OmniFlowNet~\cite{artizzu2021omniflownet} and Yuan~\etal~\cite{yuan2021360}, we use their official released codes for testing. Unfortunately, neither the code nor the dataset of LiteFlowNet360~\cite{bhandari2021revisiting} is publicly available, therefore we cannot make a fair comparison with it.
Since OmniFlowNet is an adaptive method designed for learning-based optical flow networks, we additionally upgrade its backbone from LiteFlowNet2~\cite{hui20liteflownet2} to RAFT~\cite{teed2020raft} and CSFlow~\cite{shi2022csflow} for quantitative experiments to demonstrate that our PanoFlow framework is more generic and effective. We further conduct qualitative comparisons on the public outdoor panoramic dataset OmniPhotos~\cite{bertel2020omniphotos} and our PAL-collected panoramic videos.

\subsection{Training Details}
Following previous works, we pretrain our model using the FlyingChairs~\cite{dosovitskiy2015flownet}$\rightarrow$FlyThings~\cite{mayer2016large} schedule, followed by finetuning on our FlowScape dataset. We divide FlowScape into $5,000$/$1,400$ image pairs for train/test subsets. Considering that the sunny days are the most common weather conditions, the test set of FlowScape covers sunny ($57.1\%$), cloud ($14.3\%$), fog ($14.3\%$), and rain ($14.3\%$). We train our model on an RTX 3090 GPU, implemented in PyTorch. We pretrain on FlyingChairs for $100k$ iterations with a batch size of $10$, then train for $100k$ iterations on FlyingThings3D with a batch size of $6$. Finally, we finetune on FlowScape with a batch size of $6$ for another $100k$ iterations using the weights from the pretrained model. The ablation experiment is performed with $100k$ training iterations on Chairs, and the batch size is also $10$. We time our method using an RTX 3090 GPU. The GRU iteration number is set to $12$ during training and inference. We follow RAFT~\cite{teed2020raft} for data augmentation. All experiments are with the same augmentations including occlusion augmentation~\cite{yang2019hierarchical}, random rescale, perturbing brightness, as well as contrast augmentation, saturation augmentation, and hue augmentation.

\subsection{PanoFlow on FlowScape}
\label{sec:resultsFlowScape}

We evaluate PanoFlow on the FlowScape dataset using the test split.
Results are shown in Tab.~\ref{tab:quantitative}, where we split the results based on the weather conditions. 
The best results are bolded, the second best are underlined. 
We term the method using the PanoFlow framework as PanoFlow $( \cdot )$. We denote $*$ and $**$ to distinguish models using FDA-R and FDA-E methods. C+T means that the models are trained on FlyingChairs (C) and FlyingThings (T).
F indicates methods using only FlowScape (F) train split for finetuning.
When using C+T for training, our method achieves an $11.7\%$ error reduction for RAFT, and a $12.4\%$ error reduction for CSFlow. The results of CSFlow are slightly better than RAFT, which demonstrates its better cross-dataset generalizability. 
After finetuning on FlowScape, estimating flow under PanoFlow framework can further improve the accuracy. 
Our PanoFlow (CSFlow) improves EPE from $4.47$ to $3.25$ (${\uparrow}27.3\%$).
Interestingly, FDA-R also makes it easier for the model to cope with ERP deformations when trained on the perspective dataset.
When we turn off FDA and train on FlowScape, the FDA-E models have a slightly better overall accuracy than FDA-R, but this advantage does not hold in all weather conditions.
We believe this is due to the fact that, by introducing deformed optical flow field for training process which is changed randomly in each iteration instead of being fixed, our model is able to extract robust features for computing visual similarity across different distortion modalities.

\subsection{Ablation Studies}
\label{sec:ablation}
To demonstrate the role of each core module in the proposed PanoFlow framework, we perform the ablation studies on FlowScape using the well-known RAFT structure~\cite{teed2020raft}. End-Point-Error (EPE) is used as the evaluation metric. We now describe the findings of each study.
\newpage

\begin{table}[t]
   \caption{\textbf{Quantitative results on FlowScape dataset.} \\ * denotes the model trained with FDA-R. \\ ** denotes the model trained with FDA-E.}
   \label{tab:quantitative}
   \resizebox{0.48\textwidth}{!}{
   \centering
   \renewcommand\arraystretch{1.5}{\setlength{\tabcolsep}{1mm}{\begin{tabular}{ccccccccc}
            \toprule
            \multirow{2}{*}{Training Data}& \multirow{2}{*}{Method} & \multicolumn{1}{c}{Sunny} & \multicolumn{1}{c}{Cloud} & \multicolumn{1}{c}{Fog} & \multicolumn{1}{c}{Rain} & \multicolumn{1}{c}{All (test)}  &
            \multirow{2}{*}{Diff.}
            \\
            \cline{3-7}
            & & EPE & EPE & EPE & EPE & EPE      
            \\
            \midrule
            \multirow{6}{*}{C+T} 
            & RAFT\cite{teed2020raft} & 16.57 & 11.16 & 15.04 & 17.00 & 15.64 & - \\
            & PanoFlow (RAFT)* & 14.93 & 11.25 & 13.88 & 13.36 & 14.03 & $\uparrow\textbf{10.3}\%$ \\
            & PanoFlow (RAFT)** & \underline{14.66} & \underline{11.10} & \underline{13.57} & \underline{13.38} & \underline{13.81} & $\uparrow\textbf{11.7}\%$ \\
            \cline{2-8}
            & CSFlow\cite{shi2022csflow} & 16.32 & 11.16 & 14.99 & 16.04 & 15.35 & - \\
            & PanoFlow (CSFlow)* & 14.74 & 11.18 & 13.64 & 13.42 & 13.89 & $\uparrow\textbf{9.5}\%$
            \\
            & PanoFlow (CSFlow)** & \textbf{14.27} & \textbf{10.74} & \textbf{13.03} & \textbf{13.34} & \textbf{13.45} & $\uparrow\textbf{12.4}\%$
            \\
            
            \midrule
            \multirow{6}{*}{C+T+F} 
            & RAFT\cite{teed2020raft} & 4.77 & 1.52 & 4.84 & 6.07 & 4.50 & - \\
            & PanoFlow (RAFT)* & 3.62 & \underline{1.38} & 3.60 & 4.25 & 3.39 & $\uparrow\textbf{24.7}\%$ \\
            & PanoFlow (RAFT)** & 3.58 & 1.41 & 3.63 & 4.17 & 3.36 & $\uparrow\textbf{25.3}\%$ \\
            \cline{2-8}
            & CSFlow\cite{shi2022csflow} & 4.70 & 1.46 & 4.79 & 6.24 & 4.47 & - \\
            & PanoFlow (CSFlow)* & \underline{3.56} & 1.47 & \textbf{3.56} & \textbf{3.94} & \underline{3.31} & $\uparrow\textbf{26.0}\%$ \\
            & PanoFlow (CSFlow)** & \textbf{3.46} & \textbf{1.35} & \underline{3.59} & \underline{3.98} & \textbf{3.25} & $\uparrow\textbf{27.3}\%$ \\
            
            \bottomrule
         \end{tabular}}}}
\end{table}

\begin{table}[t]
  \caption{\textbf{Ablations on Flow Distortion Augmentation.}}
  \label{tab:ablation_FDA}
  \resizebox{0.48\textwidth}{!}{
  \centering
  \renewcommand\arraystretch{1.4}{\setlength{\tabcolsep}{2.2mm}{\begin{tabular}{ccccccc}
            \toprule
            \multicolumn{2}{c}{\underline{Augmentation}} & \multirow{2}{*}{Sunny} & \multirow{2}{*}{Cloud} & \multirow{2}{*}{Fog} & \multirow{2}{*}{Rain} & \multirow{2}{*}{All (test)}                           \\   
            FDA-R & FDA-E &  &  &  &  &   
            \\
            
            \midrule
            
            - & - & 18.53 & 12.88 & 17.00 & 18.02 & 17.43\\
            \Checkmark & - & 17.86 & 12.63 & 16.60 & 17.59 & 16.89 \\
            - & \Checkmark & \textbf{16.65} & \textbf{11.67} & \textbf{15.48} & \textbf{16.50} & \textbf{15.75} \\

            \bottomrule
         \end{tabular}}}}
\label{table:abcontext}
\end{table}

\begin{table}[t]
   \caption{\textbf{Ablations on Core Components of PANOFLOW.}}
   \label{tab:ablation}
   \resizebox{0.48\textwidth}{!}{
   \centering
   \renewcommand\arraystretch{1}{\setlength{\tabcolsep}{1.5mm}{\begin{tabular}{cccccccc}
            \toprule
            \multicolumn{3}{c}{\underline{Core Components}} & \multirow{2}{*}{Sunny} & \multirow{2}{*}{Cloud} & \multirow{2}{*}{Fog} & \multirow{2}{*}{Rain} & \multirow{2}{*}{All (Test)}                           \\
            FDA-E & CFE & DCN & & & & & 
            \\
            
            \midrule
            - & - & - & 18.53 & 12.88 & 17.00 & 18.02 & 17.43 \\
            \Checkmark & - & - & 16.65 & 11.67 & 15.48 & 16.50 & 15.75 \\
            - & \Checkmark & - & 16.56 & 12.46 & 15.25 & 15.35 & 15.62 \\
            - & - & \Checkmark & 18.11 & 12.69 & 16.66 & 17.78 & 17.08 \\
            - & \Checkmark & \Checkmark & 15.93 & 12.13 & 14.69 & 15.03 & 15.08 \\
            \Checkmark & - & \Checkmark & 16.44 & 11.41 & 15.25 & 16.07 & 15.50 \\
            \Checkmark & \Checkmark & - & 14.72 & 11.28 & 13.69 & 13.84 & 13.96 \\
            \Checkmark & \Checkmark & \Checkmark & \textbf{14.55} & \textbf{11.09} & \textbf{13.57} & \textbf{13.42} & \textbf{13.75} \\
            
            \bottomrule
         \end{tabular}}}}
\end{table}

\begin{table}[t]
   \caption{\textbf{Cyclic Flow Estimation Ablation.}}
   \label{tab:CFE_ablation}
   \resizebox{0.48\textwidth}{!}{
   \centering
   \renewcommand\arraystretch{1.425}{\setlength{\tabcolsep}{0.5mm}{\begin{tabular}{cccccccc}    
            \toprule
            \multirow{2}{*}{CFE Settings}&\multicolumn{4}{c}{\underline{  \quad FlowScape (test)  \quad}}&\multirow{2}{*}{Avg.}&\multirow{2}{*}{Diff.}&\multirow{2}{*}{Latency}  
            \\
            & sunny & cloud & fog & rain & & & 
            \\    
            
            \midrule
            Baseline                & 4.77 & 1.52 & 4.84 & 6.07 & 4.50 & - & \textbf{0.10}s  \\
            Circular Convolution    & 5.72 & 2.73 & 6.02 & 7.50 & 5.59 & $\downarrow19.5\%$ & 0.11s \\
            Double Estimation       & 3.81 & 1.68 & 3.86 & 4.29 & 3.58 & $\uparrow20.4\%$ & 0.18s \\
            Half Zero Padding       & 3.82 & 1.54 & 3.74 & 4.57 & 3.59 & $\uparrow20.2\%$ & 0.24s  \\
            Half Same Padding       & 31.5 & 23.5 & 22.1 & 35.8 & 29.6 & $\downarrow558\%$ & 0.13s  \\
            Default                 & \textbf{3.58} & \textbf{1.41} & \textbf{3.63} & \textbf{4.17} & \textbf{3.36} & $\uparrow\textbf{25.3}\%$ & 0.13s  \\
        
            \bottomrule
         \end{tabular}}}}
\end{table}

\noindent\emph{\textbf{Flow Distortion Augmentation:}} We explore the role of two different flow distortion variants on the model's ability to adapt from pinhole to panoramic domains. The results are shown in Tab.~\ref{tab:ablation_FDA}. Both FDA-R and FDA-E can help models to overcome ERP deformation, which indicate that distorted optical flow is beneficial for the model to learn robust features. Although the total number of effectively supervised pixels is reduced in FDA-E, its modality is closer to ERP, thus the model using FDA-E gains an advantage.
In the following ablation experiments, we use the FDA-E model by default.

\noindent\emph{\textbf{Core Components:}} Tab.~\ref{tab:ablation} shows how performance varies as each core component (FDA-E: flow distortion augmentation in ERP format; CFE: cyclic flow estimation; DCN: deformable receptive field encoder) of our model is removed.
We can see that every component contributes to the overall performance. We find that CFE has the greatest impact on accuracy. The result is surprising, considering that this method that can be used without any retraining. This also reveals that the PanoFlow framework can easily benefit from advances in general optical flow networks. When all the key components are in place, the model performs optimally in all weathers.
In the following experiments, we use the ``full'' version of our method (last row of Tab.~\ref{tab:ablation}).

\noindent\emph{\textbf{Cyclic Flow Estimation:}} We additionally conduct an ablation study based on PanoFlow (RAFT) that has been finetuned on FlowScape to further investigate how the setting of the CFE affects the accuracy and efficiency. Tab.~\ref{tab:CFE_ablation} shows that CFE improves the performance the most in the default setting.

\emph{Circular Convolution}: In order to explore the ability of circular convolution to capture large-displacement cyclic visual similarity, we replace the convolutional layers in the model with circular convolutions, where experimental results show they do not help performance. We believe that this is because the circular convolution uses a simple padding operation to warp the image, and the introduced cyclicity is insufficient, considering that the end point of the $360^\circ$ optical flow may fall within the area $[0, \frac{W}{2}]$ outside the left and right boundaries of the panoramic image. However, most of the flow vectors are still given in the direction of the traditional optical flow, which makes it a disadvantage in $360^\circ$ flow estimation $({\downarrow}19.5\%)$. \emph{Double Estimation}: A naive idea is to swap the left and right regions directly, estimate twice and take the respective minimum values. This does improve the accuracy, but the time complexity is also doubled.
It also confuses the model during encoding the false image boundary introduced by the swap operation.
\emph{Half Zero Padding}: Based on the above observations, we naturally associate whether the another region's feature will interfere with the results of the region of interest when decoding. Thus, we try replacing half of the feature maps with empty tensors, resulting in one encoding and four decodings.
We find that it has no advantages over the default setting.
\emph{Half Same Padding}: We further replace the zero feature with same feature of the region of interest. The same features make the model face two confusing scene cues at the same time when calculating the visual similarity, which leads to terrible performance regression.
\emph{Default}: The performance improvement brought by CFE is the most significant in the default setting, and its time complexity is only modest.

\begin{figure}[t]
   \centering
   \includegraphics[width=0.85\linewidth]{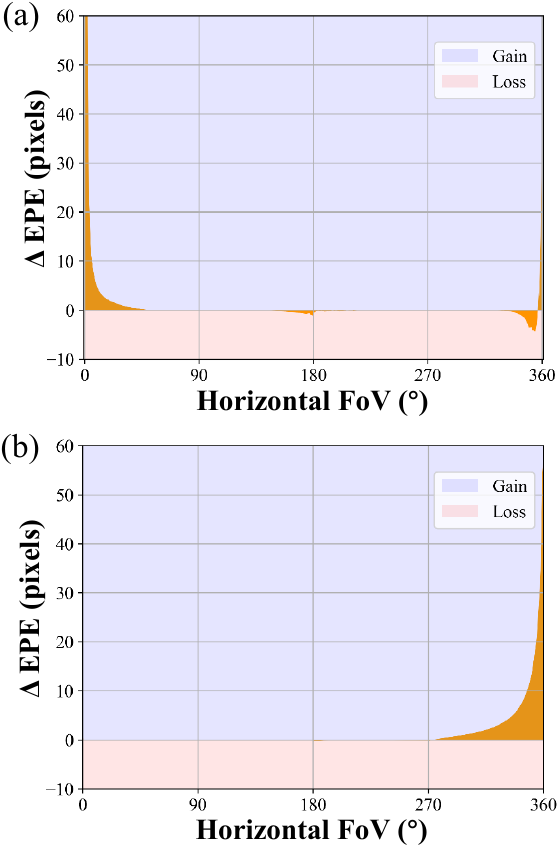}
   \caption{The distribution of the optical flow estimation EPE variation with the horizontal FoV introduced by the CFE method. (a) Statistical results on FlowScape test split, (b) Statistical results on the OmniFlowNet dataset.}
   \label{fig:deltaEPE_cyclic}
\end{figure}

We further explore the horizontal distribution of the gain introduced by CFE. As shown in Fig.~\ref{fig:deltaEPE_cyclic}, CFE improves the model's ability to cope with cross-boundary optical flow, which is an essential difficult part of panoramic flow estimation. On FlowScape, CFE seems to cause a slight degradation near the right boundary, which we believe is due to the fact that the road features of FlowScape are highly similar, causing the model to confuse the road on both sides of the boundary during cyclic inference. And our virtual collection vehicle goes straight ahead, resulting in 
less cross-boundary optical flow. On the OmniFlowNet dataset, it can be observed that the accuracy is significantly improved in the range of $270^\circ{\sim}360^\circ$ in FoV. This is reasonable because the dataset only contains forward- and rightward movements, and the crossing of boundaries generally occurs on the right side of the image. Considering real-world cases, the vehicle cannot move completely straight ahead along the lane line, and traffic accidents might occur when the vehicle turns, thus, CFE is ideally suitable for real driving scenarios.

\subsection{Comparison with the State-of-the-Art}

\noindent\emph{\textbf{Quantitative Comparison on Synthetic Data:}}
OmniFlowNet~\cite{artizzu2021omniflownet} is a state-of-the-art CNN adaption model for optical flow estimation in omnidirectional images which can be built on general CNN architectures for perspective images.
\newpage
\noindent We reproduce the OmniFlowNet using MMFlow~\cite{2021mmflow} and compare it with our model. Since OmniFlowNet is built on LiteFlowNet2~\cite{hui20liteflownet2}, which is inconsistent to our baseline, we also apply the architecture of OmniFlowNet to RAFT and CSFlow. As shown in Tab.~\ref{tab:compararison}, we evaluate the models on the OmniFlowNet dataset~\cite{artizzu2021omniflownet} with all three scenarios (\ie, CartoonTree (Cart.), Forest (Forest), LowPolyModels (Poly.)), and our FlowScape dataset. All models are trained on FlyingChairs (C) + FlyThings (T), with ``-ft'' indicating that the model was additionally fine-tuned on the FlowScape data. We also report the accuracy of the icosahedron tangent-plane panoramic flow estimation method~\cite{yuan2021360} on both datasets.

\begin{table}[!t]
\caption{\textbf{Comparison with State-of-the-art.} \\ $-ft$ denotes the model fine-tuned on FlowScape}
   \label{tab:compararison}
   \resizebox{0.49\textwidth}{!}{
   \centering
   \renewcommand\arraystretch{1.6}{\setlength{\tabcolsep}{0.5mm}{\begin{tabular}{ccccccccc}
   \toprule
    \multirow{2}{*}{Method} & \multicolumn{5}{c}{OmniFlowNet Dataset} & \multicolumn{2}{c}{ FlowScape (test) } & \multirow{2}{*}{ Latency }  \\
    & Cart. & Forest & Poly. & \multicolumn{1}{c}{Avg.} & Diff. & \multicolumn{1}{c}{Avg.} & Diff. \\
    
    \midrule
    OmniFlowNet~\cite{artizzu2021omniflownet}        & 5.37 & 8.68 & 7.32 & 7.12 & - & 22.16 & - & 0.02s\\
    Yuan~\etal~\cite{yuan2021360} & 9.13 & 14.27 & 10.22 & 11.21 & $\downarrow57.4\%$ & 20.35 & $\uparrow8.17\%$ & 10.48s\\
    OmniFlowNet (RAFT) & 4.84 & 8.70 & 6.74 & 6.76 & $\uparrow5.06\%$ & 19.61 & $\uparrow11.5\%$ & 0.43s\\
    OmniFlowNet (CSFlow)  & 4.74 & 8.66 & 6.52 & 6.64 & $\uparrow6.74\%$ & 19.47 & $\uparrow12.1\%$ & 0.44s\\
    OmniFlowNet (RAFT)-ft  & 3.55 & 7.28 & 5.28 & 5.37 & $\uparrow24.6\%$ & 14.33 & $\uparrow35.3\%$ & 0.43s\\
    OmniFlowNet (CSFlow)-ft  & 3.57 & 7.21 & 5.50 & 5.43 & $\uparrow23.7\%$ & 15.33 & $\uparrow30.8\%$ & 0.44s\\
    
    \midrule
    PanoFlow (RAFT)*     & 3.95 & 4.77 & 6.78 & 5.17 & $\uparrow27.4\%$ & 14.03 & $\uparrow36.7\%$ & 0.13s\\
    PanoFlow (RAFT)**     & 2.71 & 4.14 & 5.29 & 4.05 & $\uparrow43.1\%$ & 13.81 & $\uparrow37.7\%$ & 0.13s\\
    PanoFlow (RAFT)*-ft   & 2.31 & 3.53 & 4.91 & 3.58 & $\uparrow49.8\%$ & 3.39 & $\uparrow84.7\%$ & 0.13s\\
    PanoFlow (RAFT)**-ft   & \textbf{1.97} & \textbf{3.29} & \textbf{4.24} & \textbf{3.17} & $\uparrow55.5\%$ & \textbf{3.36} & $\uparrow84.8\%$ & 0.13s\\
    
    \midrule
    PanoFlow (CSFlow)*     & 3.81 & 4.76 & 6.92 & 5.16 & $\uparrow27.5\%$ & 13.89 & $\uparrow37.3\%$ & 0.14s\\
    PanoFlow (CSFlow)**     & 2.83 & 4.58 & 5.57 & 4.33 & $\uparrow39.2\%$ & 13.45 & $\uparrow39.3\%$ & 0.14s\\
    PanoFlow (CSFlow)*-ft   & 2.02 & \textbf{3.51} & 4.48 & 3.34 & $\uparrow53.1\%$ & 3.31 & $\uparrow85.1\%$ & 0.14s\\
    PanoFlow (CSFlow)**-ft   & \textbf{1.92} & 3.53 & \textbf{4.37} & \textbf{3.27} & $\uparrow54.1\%$ & \textbf{3.25} & $\uparrow85.3\%$ & 0.14s\\
    
    \bottomrule
    \end{tabular}}}}
    \vspace{-0.5em}
\end{table}

When the method of OmniFlowNet is applied to RAFT (OmniFlowNet~(RAFT)), the results are improved to some extent (${\uparrow}5.1\%$ on the OmniFlowNet dataset). However, PanoFlow (RAFT)** improves the performance significantly (${\uparrow}43.1\%$ on the OmniFlowNet dataset). 
After finetuning on FlowScape, the accuracy of both networks are further improved, indicating that the FlowScape dataset is effective for panoramic optical flow tasks. When training with FDA-R, PanoFlow (CSFlow)*-ft achieves better performance than PanoFlow (RAFT)*-ft on both datasets, which proves that our CSFlow structure is better at learning robust features across different distortion modalities. PanoFlow (CSFlow)**-ft ranks 1st on our FlowScape dataset ($3.25$ pixels), while PanoFlow (RAFT)**-ft gives better results on the OmniFlowNet dataset ($3.17$ pixels), which we believe is due to the domain gap between the two datasets.
On the OmniFlowNet dataset, our approach achieves a $55.5\%$ error reduction than the current state-of-the-art panoramic flow method. 
On FlowScape, PanoFlow also outperforms OmniFlowNet by a large margin, dramatically decreasing EPE from $22.16$ to $3.25$. We also perform a speed test of existing algorithms on the FlowScape dataset, in which the average time of flow computation for $100$ frames are reported. As shown in the last column of Tab.~\ref{tab:compararison}, PanoFlow yields state-of-the-art accuracy while maintaining high efficiency ($0.14$s), indicating that the proposed method achieves a better trade-off between accuracy and latency than existing methods.

\begin{figure*}[!t]
   \centering
   \includegraphics[width=1.0\linewidth]{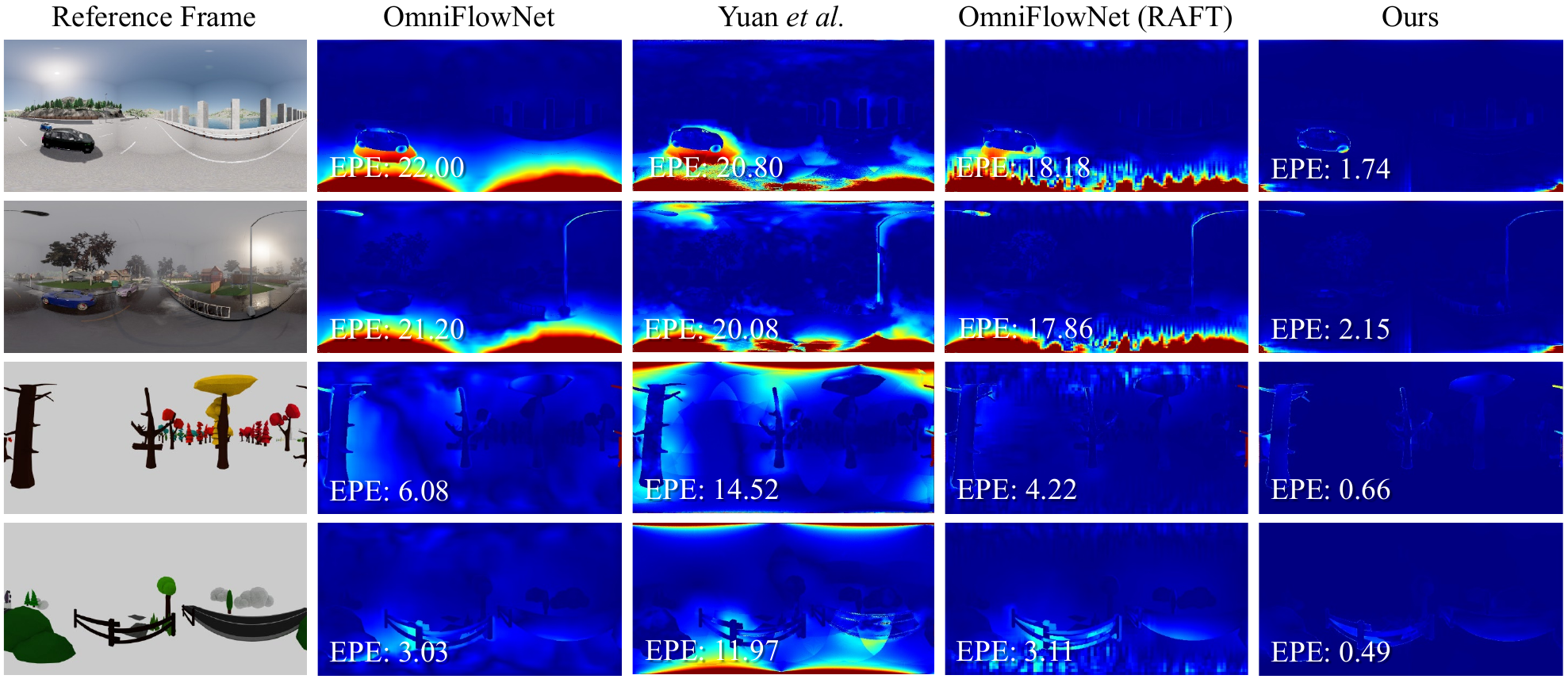}
  \vspace{-1.25em}
   \caption{Error heatmap visualizations on FlowScape test split and OmniFlowNet datasets~\cite{artizzu2021omniflownet}. PanoFlow can easily cope with the challenges introduced by image distortion in high-latitude regions and provide a clear and smooth panoramic flow field in one shot.}
   \label{fig:sync_compare}
\end{figure*}

\begin{figure*}[h]
  \centering
  \includegraphics[width=1.0\linewidth]{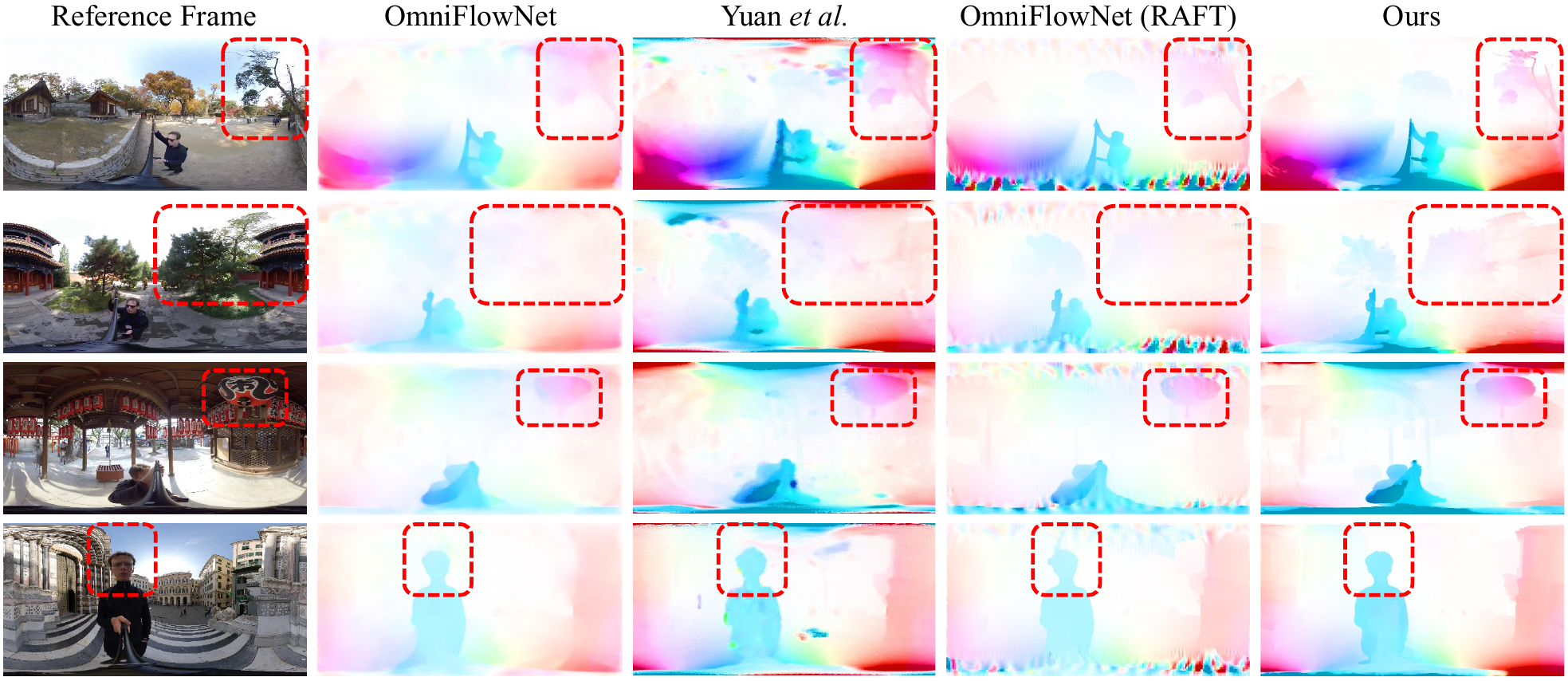}
  \vspace{-1.25em}
  \caption{Qualitative results on the OmniPhotos~\cite{bertel2020omniphotos} dataset. PanoFlow successfully generalizes from synthetic dataset to real scenes, and the panoramic flow field visualizations are clean and discriminative while well preserving the details of the image.}
  \label{fig:OP_compare}
\end{figure*}

We present the error heatmap analysis on both datasets in Fig~\ref{fig:sync_compare}. Although the convolution kernel adopts a fixed ERP deformation, OmniFlowNet~\cite{artizzu2021omniflownet} still has difficulties in dealing with high-latitude distortion. This defect is especially pronounced on FlowScape, since our dataset considers both foreground- and background panoramic optical flow, while OmniFlowNet dataset only 
gives foreground ground-truth flow. On the other hand, when the pre-trained RAFT is converted to OmniFlowNet (RAFT), we observe mosaic-like estimation errors in the high-latitude regions of the ERP, indicating the insufficient generality of the OmniFlowNet method. Since Yuan~\etal estimate the panoramic flow on both the icosahedron and the cubemap multiple times, the results tend to appear many tangent plane boundaries. Thanks to our FDA and deformable receptive field encoders, PanoFlow can easily handle high-latitude deformation of ERP. Noticeably, applying CFE produces a large improvement near the boundary, which is an essential difficult part of $360^\circ$ flow estimation (see the third row of Fig~\ref{fig:sync_compare}).

\noindent\emph{\textbf{Qualitative Comparisons in the Real World:}}
Considering that the ground truth of dense flow field in the real world is almost impossible to obtain~\cite{butler2012naturalistic,mayer2016large}, we qualitatively compare the flow on both the public OmniPhotos dataset~\cite{bertel2020omniphotos} without the ground truth and the PAL video stream collected by ourselves to verify the synthetic-to-real generalization ability of PanoFlow. As shown in Fig~\ref{fig:OP_compare}, PanoFlow gives high-quality dense optical flow in real scenes. Regarding OmniFlowNet and Yuan~\etal, they suffer from limited distortion-aware capacity, thus being not accurate enough in the real-world domain.
Although we additionally upgrade the backbone of OmniFlowNet to RAFT, its ability to capture large displacements is still insufficient compared with our method. OmniFlowNet (RAFT) also exhibits mosaic-like errors again in the real scene, proving the lack of generality of the method. Moreover, PanoFlow is able to capture adequate details of real-world images, which is evidently better than previous works.

\begin{figure*}[!t]
  \centering
  \includegraphics[width=1.0\linewidth]{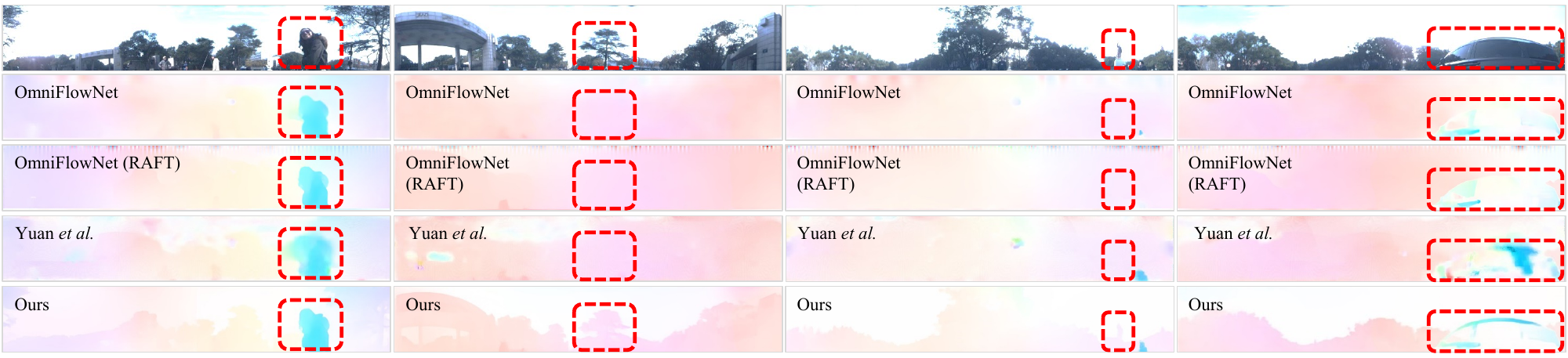}
  \caption{Qualitative comparison of existing methods in outdoor-campus $360^\circ$ image sequences that captured by our PAL camera. PanoFlow gives optical flow with clear and sharp boundaries for both foreground and background, which means stronger generalization ability for the real-world.}
  \label{fig:out_compare}
\end{figure*}

\begin{figure}[h]
  \centering
  \includegraphics[width=1.0\linewidth]{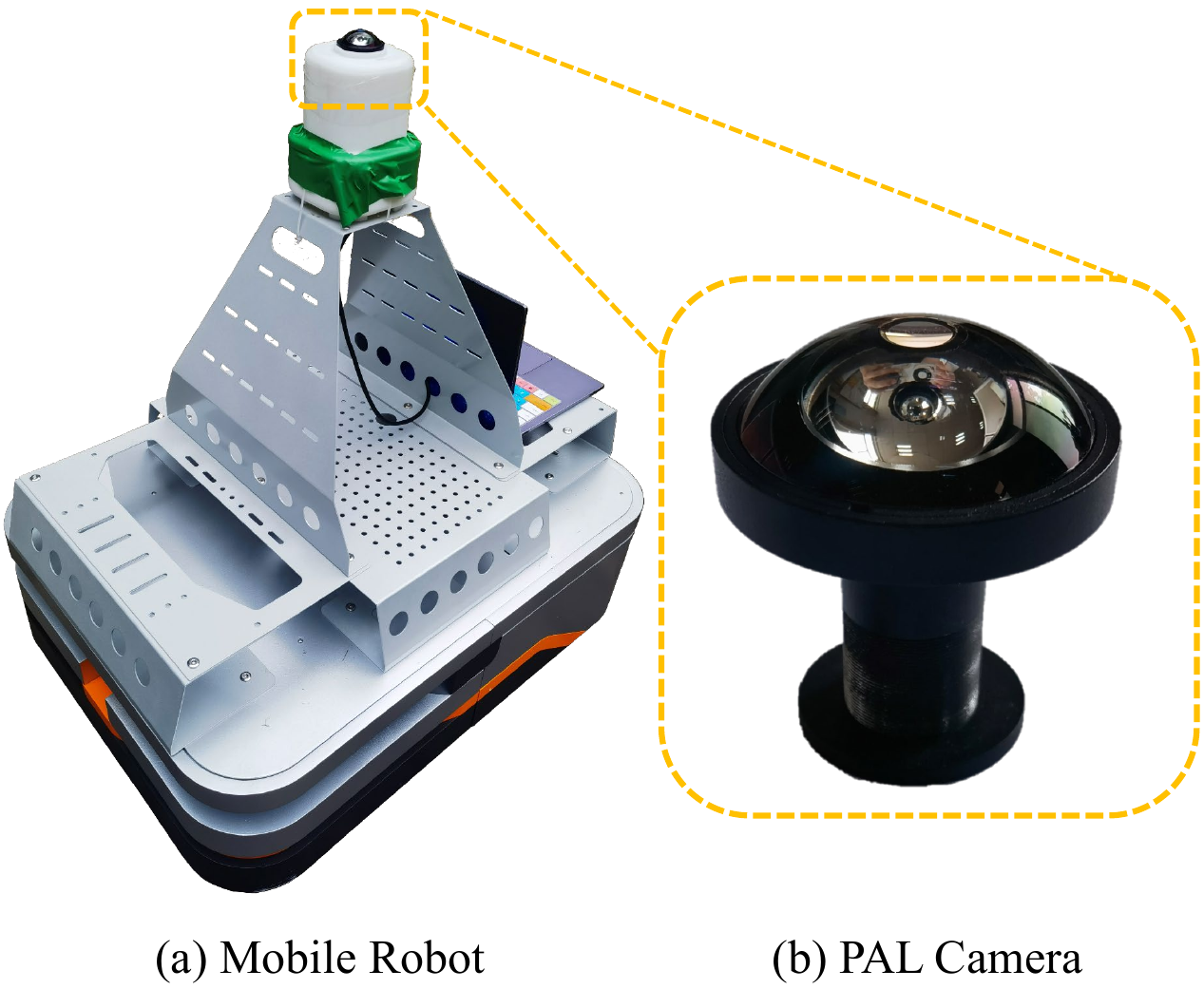}
  \vspace{-1.75em}
  \caption{(a) Our outdoor mobile robot is equipped with a Panoramic Annular Lens (PAL) camera and a laptop. (b) PAL for capturing outdoor panoramic video streams.}
  \label{fig:PAL-car}
  \vspace{-1.0em}
\end{figure}

To further investigate the practical performance of the proposed PanoFlow solution on real data, we install a panoramic annular lens (PAL) system with an FoV of $60^\circ{\times}360^\circ$ on top of a mobile robot (see Fig.~\ref{fig:PAL-car}), which navigates around the campus according to the remote control.
As shown in Fig.~\ref{fig:out_compare}, we collect panoramic videos of campus street scenes and compare our approach with the results given by OmniFlowNet~\cite{artizzu2021omniflownet} and the method from Yuan~\etal~\cite{yuan2021360}. 
Although the robot's perspective and FoV are significantly different to that of the virtual camera used in the FlowScape for training, PanoFlow still gives clear and sharp optical flow estimation. For other methods, estimating directly on PAL images will lead to epic failures. Therefore, we convert the PAL video stream to the standard ERP format with an aspect ratio of $2{:}1$ before estimation for each method except PanoFlow. This also 
reveals that these methods will face additional computational overhead when used on real panoramic shots, as their methods are only designed for complete ERP data. 
\newpage
Specifically, for pedestrian and fast-moving vehicles in the foreground of the panoramic images, 
PanoFlow does not confuse them with the motion of the background, even if they are deformed to varying degrees. Edges are blurred and indistinguishable in OmniFlowNet's background flow estimation, whereas the outlines of street scenes are still sharp and recognizable in PanoFlow's output results. Compared with the method proposed by Yuan \etal, PanoFlow gives optical flow with better continuity, and the detailed features are also well preserved.
We conclude that our method outperforms the previous state-of-the-art work for both foreground- and background motion estimations, showing excellent synthetic-to-real generalizability.
\newpage

\subsection{Efficiency Analysis}
\vspace{-0.5em}

We report the parameter counts, memory requirements during inference, inference time, and the accuracy performance as shown in Tab.~\ref{tab:timing}. Accuracy is determined by the performance on the FlowScape~(test) after training on C+T+F.
The image size is $512{\times}1024$. RAFT takes $2.67$GB memory while our approach takes $2.78$GB memory. Due to the additional global context introduced by the decoder in CSFlow, the memory consumption of PanoFlow~(CSFlow) is larger than the former. Overall, the results demonstrate that the computational overhead of PanoFlow is low, in contrast to the significant performance improvement, and is therefore suitable for intelligent vehicles to perceive surrounding temporal cues.

\begin{table}[t]
   \centering
   \caption{\textbf{Running time, parameters, and memory requirement.}}
   \label{tab:timing}
   \resizebox{0.425\textwidth}{!}{
   \centering
   \renewcommand\arraystretch{1.6}{\setlength{\tabcolsep}{2mm}{\begin{tabular}{ccccc}
            \toprule
            Method     & Parameters     & GPU Memory  & Time & $\Delta$Accuracy   \\
            
            \midrule
            RAFT~\cite{teed2020raft}                       
            & 5.3M & 2.67GB & 0.10s & -      \\
            PanoFlow (RAFT)*                       
            & 5.3M & 2.78GB & 0.13s & $\uparrow24.7\%$      \\
            PanoFlow (RAFT)**
            & 5.3M & 2.78GB & 0.13s & $\uparrow25.3\%$      \\
            CSFlow~\cite{shi2022csflow}                      
            & 5.6M & 3.42GB & 0.10s & -     \\
            PanoFlow (CSFlow)*                      
            & 5.6M & 4.04GB & 0.14s & $\uparrow26.0\%$     \\
            PanoFlow (CSFlow)**
            & 5.6M & 4.04GB & 0.14s & $\uparrow27.3\%$     \\
            
            \bottomrule
         \end{tabular}}}}
         \vspace{-1.75em}
\end{table}

\vspace{1.0em}
\begin{figure}[!t]
   \centering
   \includegraphics[width=1.0\linewidth]{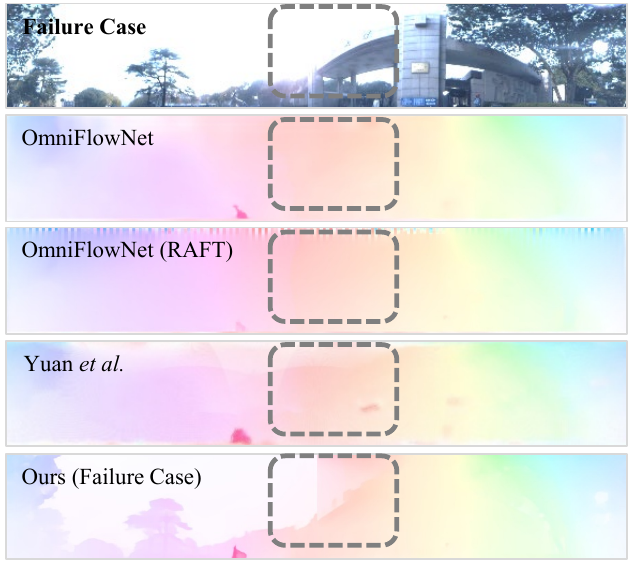}
   \caption{The failure case of PanoFlow. When there is overexposure and lack of texture near the middle of the panorama, CFE may incur difficulty distinguishing features on both sides, resulting in reduced optical flow continuity near the middle of the estimation results.}
   \label{fig:failure_case}
  \vspace{1.0em}
\end{figure}

\subsection{Failure Case Analysis}
As shown in Fig.~\ref{fig:failure_case}, when there is overexposure in the middle area of the image, the optical flow continuity on both sides will be reduced during cyclic estimation, which is reasonable because the features on both sides become difficult to distinguish and confuse the decoder of the optical flow network. Overcoming this limitation requires some form of supervision or better backbones, \eg, reasoning about panoramic semantics, reasoning about spatio-temporal features in video, or reasoning about fusion with high dynamic range sensors, such as event cameras. Future work can be dedicated to transferring our method to these approaches.

\section{Conclusion}
\input{Tex_content_red-line/conclusion}

\bibliographystyle{IEEEtran}
\bibliography{ref.bib}

\end{document}

%% file: Tex_content_red-line/abstract.tex
Optical flow estimation is a basic task in self-driving and robotics systems, which enables to temporally interpret traffic scenes. Autonomous vehicles clearly benefit from the ultra-wide Field of View (FoV) offered by $360^\circ$ panoramic sensors. However, due to the unique imaging process of panoramic cameras, models designed for pinhole images do not directly generalize satisfactorily to $360^\circ$ panoramic images. In this paper, we put forward a novel network framework——\textsc{PanoFlow}, to learn optical flow for panoramic images. To overcome the distortions introduced by equirectangular projection in panoramic transformation, we design a \emph{Flow Distortion Augmentation (FDA)} method, which contains radial flow distortion (FDA-R) or equirectangular flow distortion (FDA-E). We further look into the definition and properties of cyclic optical flow for panoramic videos, and hereby propose a \emph{Cyclic Flow Estimation (CFE)} method by leveraging the cyclicity of spherical images to infer $360^\circ$ optical flow and converting large displacement to relatively small displacement. PanoFlow is applicable to any existing flow estimation method and benefits from the progress of narrow-FoV flow estimation. In addition, we create and release a synthetic panoramic dataset \emph{FlowScape} based on CARLA to facilitate training and quantitative analysis. PanoFlow achieves state-of-the-art performance on the public OmniFlowNet and the fresh established FlowScape benchmarks. Our proposed approach reduces the End-Point-Error (EPE) on FlowScape by $27.3\%$. On OmniFlowNet, PanoFlow achieves an EPE of $3.17$ pixels, a $55.5\%$ error reduction from the best published result ($7.12$ pixels). We also qualitatively validate our method via an outdoor collection vehicle and a public real-world OmniPhotos dataset, indicating strong potential and robustness for real-world navigation applications. Code and dataset are publicly available at \href{https://github.com/MasterHow/PanoFlow}{PanoFlow}.

%% file: Tex_content_red-line/related_work.tex
\subsection{Learning-based Optical Flow Estimation}
The classical optical flow estimation approaches~\cite{horn1981determining,lucas1981iterative} use variational approaches to minimize energy based on brightness constancy and spatial smoothness.
Since the advent of FlowNet~\cite{dosovitskiy2015flownet}, some other works based on Convolutional Neural Netowrks (CNNs)~\cite{tran2016deep,ahmadi2016unsupervised,wulff2015efficient,zhang2021separable,jiang2021learning,zhao2022global_matching_overlapping_attention,bai2022deep_equilibrium,luo2022learning_kernel_patch_attention,zheng2022dip,sui2022craft,xu2022gmflow,jeong2022imposing_consistency} have appeared.
Besides, there are also some self-supervised approaches~\cite{liu2019selflow,tung2017self} to learn optical flow with occlusions.
Most of these methods are normally designed to work with pinhole cameras capturing a limited imaging angle.

FlowNet~\cite{dosovitskiy2015flownet} first treats optical flow estimation as a learning problem.
In order to further improve the accuracy of optical flow, FlowNet2.0~\cite{ilg2017flownet} introduces image warping between multiple cascaded FlowNets. Due to the large model size of FlowNet2.0~\cite{ilg2017flownet}, many methods have been proposed to simultaneously improve the optical flow accuracy and reduce the model size.
Among them, PWC-Net~\cite{sun2018pwc} combines classical optical flow estimation principles including pyramid processing, image warping, and cost volumes with learning. LiteFlowNet2~\cite{hui20liteflownet2} draws on the idea of data fidelity and regularization in the classical variational optical flow method. RAFT~\cite{teed2020raft} iteratively update optical flow fields using multi-scale 4D correlation volumes. To better apply optical flow estimation to autonomous driving systems, CSFlow~\cite{shi2022csflow} proposes a new optical flow deep network architecture composed of Cross Strip Correlation module (CSC) and Correlation Regression Initialization module (CRI). 
Moreover,
FlowFormer~\cite{huang2022flowformer} replaces the CNN-based backbone in the RAFT architecture with a transformer-based backbone, which further improves the accuracy of optical flow estimation while increasing the number of parameters by three times. In contrast, PanoFlow is a panoramic optical flow framework that can be adapted to any optical flow network with an encoder-decoder architecture.

\subsection{Optical Flow Estimation beyond the FoV}
With the arrival on the market of the increasingly affordable, portable, and accurate panoramic cameras, $360^\circ$ flow estimation is in urgent need, that can provide a wide-FoV temporal understanding, for which some methods based on deep learning are developed. LiteFlowNet360~\cite{bhandari2021revisiting} is designed as a domain adaptation framework to cope with inherent distortion in $360^\circ$ videos caused by the sphere-to-plane projection. They employ incremental transformation of convolutional layers in feature pyramid networks to reduce network growth size and computational costs combining data augmentation and self-supervised learning with target-domain $360^{\circ}$ videos.
OmniFlowNet~\cite{artizzu2021omniflownet} is built on a CNN model that specializes in perspective images and then applied to omnidirectional ones without training on new datasets, whose convolution operation is unified with equirectangular projection, outperforming the original network.
The projection from the $360^\circ$ image to the ERP image is a nonlinear mapping, and the distortion caused by this will affect the $360^\circ$ optical flow estimation, thus Yuan~\textit{et al.}~\cite{yuan2021360} propose a $360^\circ$ optical flow estimation method based on tangent images, including dozens of estimations and refinements on both icosahedron and cubemap panoramas. 
Overall, the existing learning-based panoramic flow methods adopt a fixed projection paradigm at the model level to deal with ERP distortions. Considering the local bias behavior of CNNs, this will reduce the model's ability to model potential visual cues and result in unsatisfactory performance. On the other hand, estimation and refinement on tangent planes introduce additional computational costs, leading to limited inference speed. Recently, a concurrent work~\cite{bhandari2022learning} also explores the $360^\circ$ optical flow via a siamese representation learning scheme with
carefully designed losses and rotational augmentations to adopt existing flow networks.
Differing from these works, we tackle image distortions and object deformations that appear across the entire $360^\circ$ scenes and leverage the cyclicity of consecutive omnidirectional data for enhancing panoramic optical flow estimation.

\subsection{Optical Flow and Panoramic Perception Datasets}

Panoramic datasets are needed in a wide variety of application areas, including depth estimation~\cite{im2016all,jiang2021unifuse,sun2021hohonet},
scene segmentation~\cite{zhang2022bending,deng2019restricted,jaus2022panoramic_panoptic_segmentation}, and optical flow estimation~\cite{artizzu2021omniflownet,yuan2021360,sekkat2022synwoodscape}.
Stanford2D3D~\cite{stanford2d3d} is a large-scale indoor spaces dataset that consists of both regular and panoramic data  with instance-level semantic annotations.
The 360D dataset~\cite{zioulis2018omnidepth} reuses released large-scale 3D datasets and re-purposes them to $360^\circ$ via rendering for dense depth estimation.
PASS~\cite{yang2019pass} presents a panoramic annular semantic segmentation framework with an associated dataset for credible evaluation.
DensePASS~\cite{zhang2021transfer} introduces a dataset with both labeled and unlabeled $360^\circ$ images for benchmarking panoramic semantic segmentation from a perspective of unsupervised domain adaptation. 
KITTI-360~\cite{liao2021kitti360} is collected with perspective stereo cameras, a pair of fisheye cameras, and a laser scanning unit for enabling $360^\circ$ perception.
WoodScape~\cite{yogamani2019woodscape} comprises of multiple surround-view fisheye cameras and multiple tasks like segmentation and soiling detection.
The OmniScape dataset~\cite{sekkat2020omniscape} includes semantic segmentation, depth map, intrinsic parameters of the cameras, and the dynamic parameters of the motorcycle.
The Waymo Open dataset~\cite{mei2022waymo_open_dataset} is a labeled panoramic video dataset for panoptic image segmentation.

Aiming at improving the accuracy of optical flow estimation, OmniFlow~\cite{seidel2021omniflow} is a synthetic omnidirectional human optical flow dataset with images of household activities with a FoV of $180^\circ$. OmniFlowNet~\cite{artizzu2021omniflownet} renders a test set of panoramic optical flow only for validation, using simple geometric models based on Blender. 
Replica360~\cite{yuan2021360} implements an ERP camera model for the Replica rendering pipeline~\cite{straub2019replica} and contains ground-truth optical flow in the equirectangular format for validation.
SynWoodScape~\cite{sekkat2022synwoodscape} is a synthetic fisheye surround-view dataset with ground truth for pixel-wise optical flow and depth estimation.
OmniPhotos~\cite{bertel2020omniphotos} is a fast $360^\circ$ panoramic VR photography method with an released outdoor dataset, but it cannot obtain the ground truth of optical flow.
We note that, until now, there is neither a dataset for omnidirectional images targeting at outdoor complex street scenes, nor a dataset covers $360^\circ$ that can be used for training and evaluation. The present paper seeks to fill this gap, by proposing a virtual environment, in which one car with panoramic camera drives under the assumption that pedestrians and vehicles move according to traffic rules. Tab.~\ref{tab:compare_data} relates and summarizes current panoramic datasets that contain ground-truth optical flow. A detailed analysis of the ground-truth quality of optical flow will be unfolded in Sec.~\ref{sec:FlowScape}.

%% file: Tex_content_red-line/conclusion.tex
In this paper, we proposed PanoFlow, a flexible framework for estimating $360^\circ$ optical flow using flow distortion augmentation, cyclic flow estimation, and deformable receptive filed encoder. We also proposed FlowScape, a publicly available synthetic panoramic optical flow dataset, which can be used for training and evaluation. We have proved through a large number of quantitative experiments that our PanoFlow is compatible with any optical flow methods of an encoder-decoder structure, which significantly improves the accuracy of panoramic flow estimation while ensuring computational efficiency. 
PanoFlow achieves state-of-the-art performance on both public OmniFlowNet dataset and our FlowScape. PanoFlow also demonstrates strong synthetic-to-real generalizability in the real world, giving high-quality panoramic flow fields for both foreground and background.
We look forward to further exploring the adaptability of the PanoFlow framework for other downstream panoramic tasks.
\newpage
In the future, we aim to explore other panoramic scene understanding tasks, such as the fusion of panoramic camera and LiDAR sensor for an entire and complete semantic and temporal surrounding perception. Furthermore, we plan to exploit synthetic data to study robust scene perception under corner cases such as risky driving and sensor failures to alleviate the long-tail problem in autonomous driving. We also have the intention to look into 3D scene flow estimation based on panoramic cameras. In addition to panoramic cameras with ultra-wide FoV, we are also interested in exploring optical flow estimation for event cameras with ultra-high dynamic range.